\documentclass[sigconf]{acmart}
\newcommand{\ourwork}{MPGraph}

\newcommand\blfootnote[1]{%
  \begingroup
  \renewcommand\thefootnote{}\footnote{#1}%
  \addtocounter{footnote}{-1}%
  \endgroup
}

\AtBeginDocument{%
  }

\copyrightyear{2023}
\acmYear{2023}
\setcopyright{rightsretained}
\acmConference[SC '23]{The International Conference for High Performance Computing, Networking, Storage and Analysis}{November 12--17, 2023}{Denver, CO, USA}
\acmBooktitle{The International Conference for High Performance Computing, Networking, Storage and Analysis (SC '23), November 12--17, 2023, Denver, CO, USA}
\acmDOI{10.1145/3581784.3607043}
\acmISBN{979-8-4007-0109-2/23/11}

\usepackage{paralist}
\usepackage{enumitem}
\usepackage{subfig}
\usepackage{algorithm}
\usepackage[noend]{algpseudocode}
\usepackage{url}
\begin{document}


 \title[Domain Specific ML Prefetcher for Accelerating Graph Analytics]{Phases, Modalities, Spatial and Temporal Locality:\\Domain Specific ML Prefetcher for Accelerating Graph Analytics}


\author{Pengmiao Zhang}
\orcid{0000-0002-5411-3305}
\affiliation{%
  \institution{University of Southern California}
  \city{Los Angeles}
    \state{California}
  \country{USA}
  \postcode{90089}
}
\email{pengmiao@usc.edu}

\author{Rajgopal Kannan}
\affiliation{%
  \institution{DEVCOM Army Research Lab}
  \city{Los Angeles}
  \state{California}
  \country{USA}
  \postcode{90094}
 }
\email{rajgopal.kannan.civ@army.mil}
 
\author{Viktor K. Prasanna}
\affiliation{%
  \institution{University of Southern California}
  \city{Los Angeles}
  \state{California}
  \country{USA}
  \postcode{90089}
}
\email{prasanna@usc.edu}

\renewcommand{\shortauthors}{Zhang et al.}

\begin{abstract}
{\color{black}

Memory performance is a key bottleneck in accelerating graph analytics. Existing Machine Learning (ML) prefetchers encounter challenges with phase transitions and irregular memory accesses in graph processing. We propose MPGraph, an \underline{M}L-based \underline{P}refetcher for \underline{Graph} analytics using domain specific models. MPGraph introduces three novel optimizations: soft detection of phase transitions, phase-specific multi-modality models for access delta and page predictions, and chain spatio-temporal prefetching (CSTP) for prefetch control.}

Our transition detector achieves 34.17–82.15\% higher precision compared with Kolmogorov–Smirnov Windowing and decision tree. Our predictors achieve 6.80–16.02\% higher F1-score for delta and 11.68–15.41\% higher accuracy-at-10 for page prediction compared with LSTM and vanilla attention models. Using CSTP, MPGraph achieves 12.52–21.23\% IPC improvement, outperforming state-of-the-art non-ML prefetcher BO by 7.58–12.03\% and ML-based prefetchers Voyager and TransFetch by 3.27–4.58\%. 
{\color{black}
For practical implementation, we compress the prediction models to reduce the storage and latency overhead. MPGraph with the compressed models still shows significantly superior accuracy and coverage compared to BO, with 3.58\% IPC improvement.
}

\end{abstract}





\maketitle
\begin{sloppypar}

\section{Introduction}

\blfootnote{{\color{black}Distribution Statement A: Approved for public release. Distribution is unlimited.}}

Graph analytics is widely used in scientific and engineering fields to analyze complex structural relationships in graphs.
However, the enormous size of Big Data graphs and attendant complexity of the analytics algorithms ~\cite{sundaram2015graphmat} often result in underperformance due to inefficient memory utilization (low data reuse, high cache miss rates, etc.)~\cite{basak2019analysis}. 
While many effective graph processing frameworks for improving graph analytics performance have been proposed~\cite{lakhotia2020gpop,xstream,gonzalez2012powergraph,dean2008mapreduce,ligra, graphmat, galois,polymer}, there still remains ample scope for ameliorating memory performance issues.

Data prefetching is the process of proactively fetching data into the memory cache before the data is requested.
There are several options for incorporating prefetching within graph analytics applications.
Traditional rule-based prefetchers
~\cite{8514366,4798239,byna2009taxonomy,somogyi2006spatial,joseph1997prefetching,michaud2016best,shevgoor2015efficiently,jain2013linearizing,lu2020apac,yu2015imp}, for example, 
offer the simplicity of easy hardware implementation along with some performance acceleration~\cite{vander1997caches}.  
However, these rule-based prefetchers have limited adaptability -- they cannot handle the complex memory access patterns seen in  
graph analytics. 
In contrast, Machine Learning (ML) algorithms for memory access prediction and prefetching show high adaptability and generalizability~\cite{zhang2021c,shi2021hierarchical,zhang2022fine,srivastava2019predicting,hashemi2018learning}, thereby offering a promising alternative for boosting performance. Further, their superior prediction accuracy on complex sequences and patterns makes them an attractive design choice for graph analytics acceleration.

Given the importance of the graph analytics domain, we believe there is a need to develop dedicated high-performance ML prefetchers for such applications. To the best of our knowledge, there do not exist such domain-specific ML prefetchers (for graph analytics). Developing such a prefetcher presents multiple unique challenges. 1) The memory access patterns vary across different graph processing phases~\cite{lakhotia2020gpop,xstream,pcpm}, making it difficult to train a general ML model that performs well across all phases~\cite{french1999catastrophic,kirkpatrick2017overcoming}. 2) Parallel executions under multi-core systems introduce randomness and irregularity~\cite{ma2014memory}, decreasing pattern matches in temporal prefetchers, which typically rely on recurring patterns~\cite{jain2013linearizing,shi2021hierarchical}. 3) Processing of connected nodes stored on multiple pages causes wide-range page jumps, making spatial prefetchers that predict within a page ~\cite{michaud2016best,zhang2022fine} less effective. Existing ML-based prefetchers~\cite{hashemi2018learning,peled2018neural,shi2021hierarchical,zhang2022fine} model the prefetching process as a simple sequence prediction problem. Such a modeling approach integrates little or no domain knowledge of applications and does not address the above challenges. 

In this paper, we present~\ourwork, the first ever \underline{M}L-based \underline{P}refetcher for \underline{Graph} analytics. Uniquely, \ourwork~ is built over a collection of domain specific (DS) ML models, which capture the domain specific context of both {\it architecture} and {\it computation} in graph analytics. Specifically, ~\ourwork~is optimized along three DS dimensions: \textit{phases}, \textit{modalities}, and \textit{locality}.

\textbf{Optimizations for phases.} 
Graph processing applications often exhibit different memory access patterns in different phases. To improve prediction accuracy, we propose phase-specific models. However, detecting phase transitions in a timely and reliable manner is challenging. We address false detections caused by impulse pattern shifts and present two methods for phase transition detection based on whether phase labels are accessible or not. If the phase labels are inaccessible, we develop an unsupervised model called Soft-KSWIN, which is a variant of the windowing Kolmogorov–Smirnov test (KSWIN). If the phase labels are accessible, we train a decision tree classifier offline by supervised learning. Both methods use a soft detection scheme to avoid false positive in detections.

\textbf{Optimizations for modalities.} 
Modality refers to the way of describing an event or experience~\cite{baltruvsaitis2018multimodal}. We propose a novel approach to model memory access prediction based on modality. We treat the program counter (PC) sequence as a distinct modality, which describes the memory accesses from the instruction perspective. We introduce AMMA, an \underline{A}ttention-based network structure using \underline{M}ulti-\underline{M}odality \underline{A}ttention fusion, which combines the address input and PC input using an attention mechanism~\cite{nagrani2021attention,jewitt2016introducing}. The attention mechanism~\cite{vaswani2017attention} has shown success in prefetching due to high adaptability and parallelizability. Our AMMA model outperforms existing methods on various benchmarks and datasets.

\textbf{Optimizations for locality.} 
To address the limitations of prefetchers relying only on spatial or temporal locality, we propose to exploit both by combining memory access page prediction with delta prediction. We predict a future memory page based on temporal locality and multiple deltas within a page based on spatial locality. Then we develop a novel \underline{C}hain \underline{S}patio-\underline{T}emporal \underline{P}refetching (CSTP) strategy, which uses the predicted page as the base for delta prediction and forms a chain of prefetch requests.

We conduct a comprehensive evaluation of our approach using three state-of-the-art graph processing frameworks: GPOP~\cite{lakhotia2020gpop}, X-Stream~\cite{xstream}, and PowerGraph~\cite{gonzalez2012powergraph}. These frameworks adopt a synchronous computation model, where each phase of the graph algorithm is followed by a global barrier synchronization. We use a diverse set of real-world and synthetic graphs as input data for our experiments. We use ChampSim~\cite{ChampSim} to generate memory access traces and simulate the physical memory behavior of our approach.

While our primary focus in this paper is improving prefetching accuracy, we also develop several techniques for reducing implementation complexity. 
While ML prefetchers are generally more complex than rule-based ones, such designs will become increasingly practical as hardware efficiency evolves (for example, efficient ML model parallelization).
Thus, we believe ML graph analytics prefetchers are needed even at (currently) higher implementation costs due to the promise of superior performance. Notwithstanding this, we develop several techniques for optimizing our models for practical implementation. 1) To reduce model storage, we compress the models based on binary encoding, knowledge distillation, and quantization. 2) To reduce inference latency, we analyze the critical path under parallel implementation and apply distance prefetching to hide the latency. Even our most compressed model, with drastically reduced storage/latency outperforms the best performing non-ML prefetcher and can thus be integrated into any state-of-the-art graph analytics framework as the preferred prefetcher.

We summarize our contributions below:


\begin{itemize}
    \item A methodology for developing domain specific (DS) ML models for prefetching. We analyze features that capture the context of the architecture and computation, then illustrate their application in developing a prefetcher for graph analytics.

    \item \ourwork, the first ever dedicated  ML prefetcher for graph analytics built over DS ML models optimized for phases, modalities, and locality.
    
    \item Phase transition detectors using a soft detection scheme. Our approach achieves 34.17--82.15\% higher detection precision than the KSWIN and decision tree baselines.

    \item AMMA, a multi-modality attention-based network for memory access prediction. It outperforms LSTM and vanilla attention by 6.80--16.02\% w.r.t. F1-score for delta prediction and by 11.68--15.41\% w.r.t. accuracy-at-10 for page prediction.
    
    \item CSTP, a prefetching strategy that utilizes spatial delta predictions and temporal page predictions.~\ourwork~using CSTP outperforms the widely-used rule-based prefetcher BO by 7.58--12.03\%, state-of-the-art ML-based prefetcher Voyager by 3.27--4.42, and TransFetch by 3.73--4.58\%.

    \item Optimizations towards practical prefetcher. We develop techniques for reducing model storage overhead and inference latency.~\ourwork~using $87\times$ compressed models with 200 cycles latency introduced in simulation outperforms the best performing non-ML prefetcher (Best Offset \cite{michaud2016best}) by 3.58\%. 
    
\end{itemize}

\section{Related Work}

\subsection{State-of-the-Art Data Prefetchers}
\noindent\textbf{Rule-based Prefetchers.} Traditional prefetchers learn from pre-defined rules. For example, spatial prefetcher BO~\cite{michaud2016best} and VLDP~\cite{shevgoor2015efficiently} learn from history access page offsets or deltas and predict future accesses within a spatial region. Temporal prefetchers like Irregular Stream Buffer (ISB)~\cite{jain2013linearizing} and Domino~\cite{bakhshalipour2018domino} predict temporally correlated memory accesses by recording and replaying the history access sequences. Indirect prefetcher IMP~\cite{yu2015imp} detects indirect accesses based on index-address pairs. These approaches use heuristic rules that cannot adapt to complex graph analytics memory access patterns. Recently, rule-based
prefetchers (DROPLET~\cite{basak2019analysis}, RnR~\cite{zhang2020rnr}, and Prodigy~\cite{talati2021prodigy}) were developed specifically for graph analytics. However, they require extra support from input data, programmer notation, or compiler to exploit graph analytics properties. In contrast, our work uses the same interfaces with traditional hardware prefetchers. Therefore, we use BO and ISB as rule-based baselines that are compatible with our architecture.

\noindent\textbf{ML-based Prefetchers.} 
Machine Learning (ML) has become a key technique for memory access prediction and data prefetching. Various approaches have been proposed, including logistic regression and decision trees for pattern classification~\cite{Rahman2015-xk}, reinforcement learning for context-based prefetching~\cite{peled2015semantic}, and Long Short-Term Memory (LSTM) networks to predict memory accesses ~\cite{hashemi2018learning, peled2018neural,zeng2017long,braun2019understanding, zhang2020raop,zhang2021c}. Hashemi et al. used LSTM to learn memory access delta patterns~\cite{hashemi2018learning}. Shi et al. proposed Voyager~\cite{shi2021hierarchical}, which predicts both page sequence and page offsets using two LSTM models along with a dot-product attention mechanism. Zhang et al. developed TransFetch~\cite{zhang2022fine}, an attention-based network using fine-grained address segmentation as input and achieves state-of-the-art prefetching performance. Although useful, these general models may not be the most efficient solution for specific domains such as graph analytics. Recently, ML algorithms were applied to predicting memory accesses for graph analytics~\cite{zhang2022sharp,zhang2022a2p}. However, they do not develop the models based on graph analytics properties and do not integrate the models into prefetchers.
To the best of our knowledge, our work is the first ML-based prefetcher developed specifically for the graph analytics domain, achieving superior prediction and prefetching performance.

\subsection{Incorporating Domain Knowledge into ML}

Domain knowledge have been integrated into ML models in various areas~\cite{dash2022review}, such as climate modeling~\cite{kashinath2021physics}, turbulence modeling~\cite{wu2018physics}, fire engineering~\cite{naser2021mechanistically}, earth science~\cite{karimpouli2020physics}, chemistry~\cite{liu2019predicting}, etc. Murdock et al. demonstrate the effectiveness of domain knowledge in improving ML model predictive ability~\cite{murdock2020domain}. Several notions related to domain specific ML has been explored in recent literature.  
Theory-guided data science~\cite{karpatne2017theory} proposes a new paradigm that is gaining prominence in scientific disciplines when designing data science models. Informed machine learning~\cite{von2019informed} integrates prior knowledge into the machine learning pipeline by using it as an independent input source. Physics-based machine learning~\cite{karniadakis2021physics,willard2020integrating} incorporates physical properties into model training by introducing physics-guided loss functions and initialization. 

In this work, we propose using domain specific ML in the context of memory address prediction and prefetching. Some existing ML-based prefetchers, 
though not designed specifically for graph analytics
incorporate domain knowledge 
into their models. 
For example, TransFetch~\cite{zhang2022fine} 
(memory address configuration), Voyager\cite{shi2021hierarchical} 
(program counters) and ReSemble~\cite{zhang2022resemble} (spatio-temporal localities for ensemble prefetching). We generalize this process by analyzing the architecture context of the target hardware and the computation context of the target applications to guide our modeling.

\section{Domain Specific ML for Prefetching}

\label{sec:motiv}

\noindent Previous ML-based prefetchers used general models with limited domain specific optimizations for particular classes of applications. We propose developing DS ML models for prefetchers that incorporate context features based on domain knowledge to achieve higher prefetching performance.

\subsection{Domain Specific Features for Prefetching}

By exploiting domain specific features from the context of architecture and computation for a specific domain application, we can integrate domain knowledge into ML models for prefetching.

Example features from the context of the architecture include:
\begin{itemize}
    \item \textbf{Platform:} the target architecture on which the computation is executed, such as multi-core platform, multi-CPU cluster, heterogeneous architecture, etc.
    \item \textbf{Memory hierarchy:} the parameters for each level of cache, main memory, persistent memory~\cite{yang2020empirical}, flash memory, etc.
    \item \textbf{Memory address configuration:} the bit length of a memory address, the cache line size, the page size, etc.
\end{itemize}

Example features from the context of the computation include:
\begin{itemize}
    \item \textbf{Computing paradigm:} A programming model that a particular application follows, such as MapReduce~\cite{dean2008mapreduce}, Scatter-Gather~\cite{lakhotia2020gpop,xstream}, and GAS~\cite{gonzalez2012powergraph} in graph analytics.

    \item \textbf{Phase:} a step or a super-step in the domain computation, such as the local computation step and value communication steps in distributed computing paradigm.

    \item \textbf{Modality:} a mode, a set of features, to describe the computation. For memory accesses, modalities include program counters, memory addresses, thread IDs, etc.

    \item \textbf{Locality:} the data access patterns in a computation, e.g., the spatial locality and the temporal locality.

    \item \textbf{Thread:} a unit of execution within a process. Multiple threads can run in parallel in a multi-core CPU and share the same memory space, challenging memory access prediction.
    
    \item \textbf{Coordination:} the way a parallel computing application defines how multiple cores or nodes of a system work together, including synchronous and asynchronous coordination.

\end{itemize}

\subsection{Developing Domain Specific ML Models for Graph Analytics Prefetcher}

To build a high-performance ML-based prefetcher for graph analytics, we propose developing domain specific ML models -- 
incorporating the domain specific features into the model design.

\begin{figure}[ht]
  \centering
  \includegraphics[width=1\linewidth]{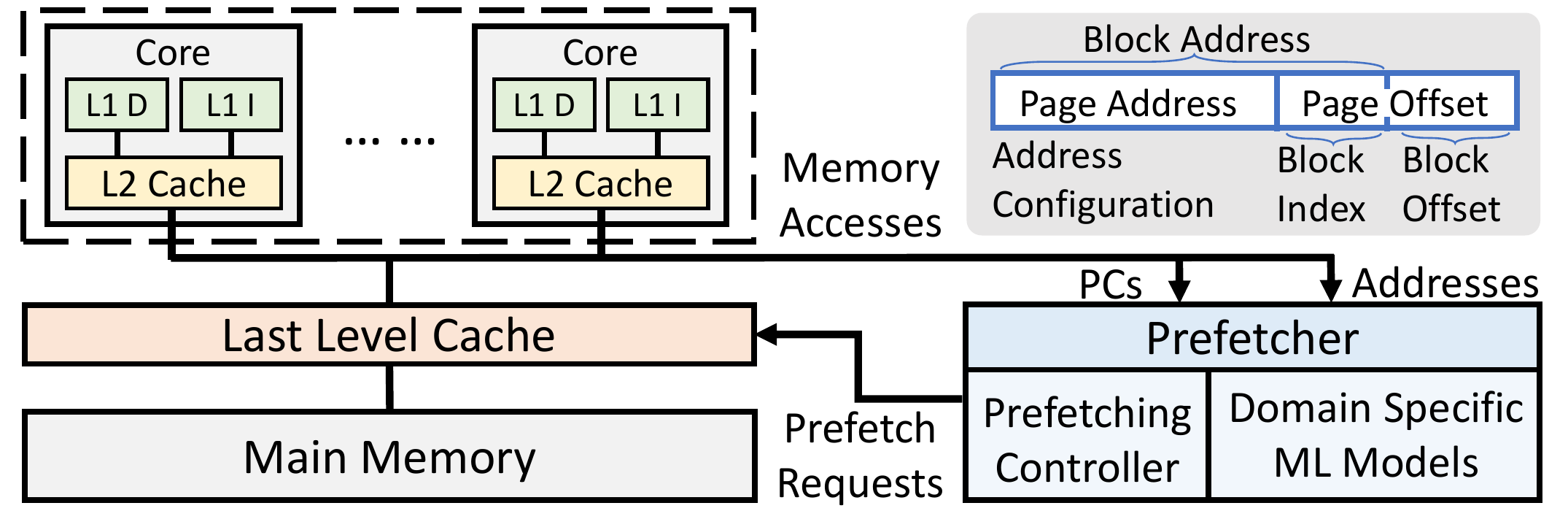}
  \caption{Multi-core shared memory architecture with ML-based LLC prefetcher.}
  \label{fig:target}
\end{figure}

In the context of architecture, our target platform is a multi-core shared memory architecture, as depicted in Figure~\ref{fig:target}, which can serve as a node of HPC. The memory hierarchy consists of private L1 caches (including data cache L1D and instruction cache L1I), private L2 caches, a shared last-level cache (LLC), and a shared main memory. This results in LLC data requests from interleaved instructions from different cores. 
Memory subsystem directly impacts the performance of the HPC system~\cite{peng2021holistic,schneider2017fast}.
Our prefetcher leverages domain specific models to predict memory accesses. We take into account the memory address configuration of the system. The prefetching is at the block level and the virtual-to-physical address translation is at the page level. Based on the configuration, we train page and block index prediction models for data prefetching.

In the context of computation, many graph processing frameworks using various computing paradigms have been developed with exceptional performance~\cite{lakhotia2020gpop,gonzalez2012powergraph,xstream,dean2008mapreduce,zhou2023interactive}. We target graph analytics frameworks with iterative barrier-synchronized phases. 
{
\color{black}
In these frameworks, the phases are defined in software and all the applications implemented using the framework follow the same programming paradigm. Thus, a domain specific ML prefetcher design is applicable to all applications developed using the framework.
We use the GPOP framework~\cite{lakhotia2020gpop} to illustrate the incorporation of domain specific features for developing domain specific ML models. GPOP is a graph processing framework based on the Scatter-Gather paradigm, as outlined in Algorithm~\ref{alg:sg}.
}
It consists of two phases: Scatter, which propagates the current value of a vertex to its neighboring vertices along edges, and Gather, which accumulates values from neighboring vertices to update the value of a vertex. Specifically, we incorporate domain specific features phase, modality, and locality into our model development.

\begin{algorithm}[t]
  \caption{{\color{black}Scatter-Gather Paradigm with Synchronized Phases}}
  \label{alg:sg}
  \begin{algorithmic}[1]
     	\Procedure{Scatter}{vertex $v$}
		\State Propagate the value of $v$ to neighbors along edge 
		\EndProcedure
		\Procedure{Gather}{vertex $v$}
		\State Accumulate values from neighbors to update $v$
		\EndProcedure
		\While{not done}
		\ForAll{vertex $v$ that need to scatter updates}
		    \State \Call{Scatter}{$v$} \Comment{Scatter phase}
		\EndFor
		\State{$\_\_synchronize()\_\_$} 
		\ForAll{vertex $v$ that have updates}
		    \State \Call{Gather}{$v$} \Comment{Gather phase}
		\EndFor
		\State{$\_\_synchronize()\_\_$} 
		\EndWhile
  \end{algorithmic} 
\end{algorithm}

\begin{figure}[t]
    \centering
  \subfloat[PCA for memory accesses.\label{fig:pca1}]{%
        \includegraphics[width=0.5\linewidth]{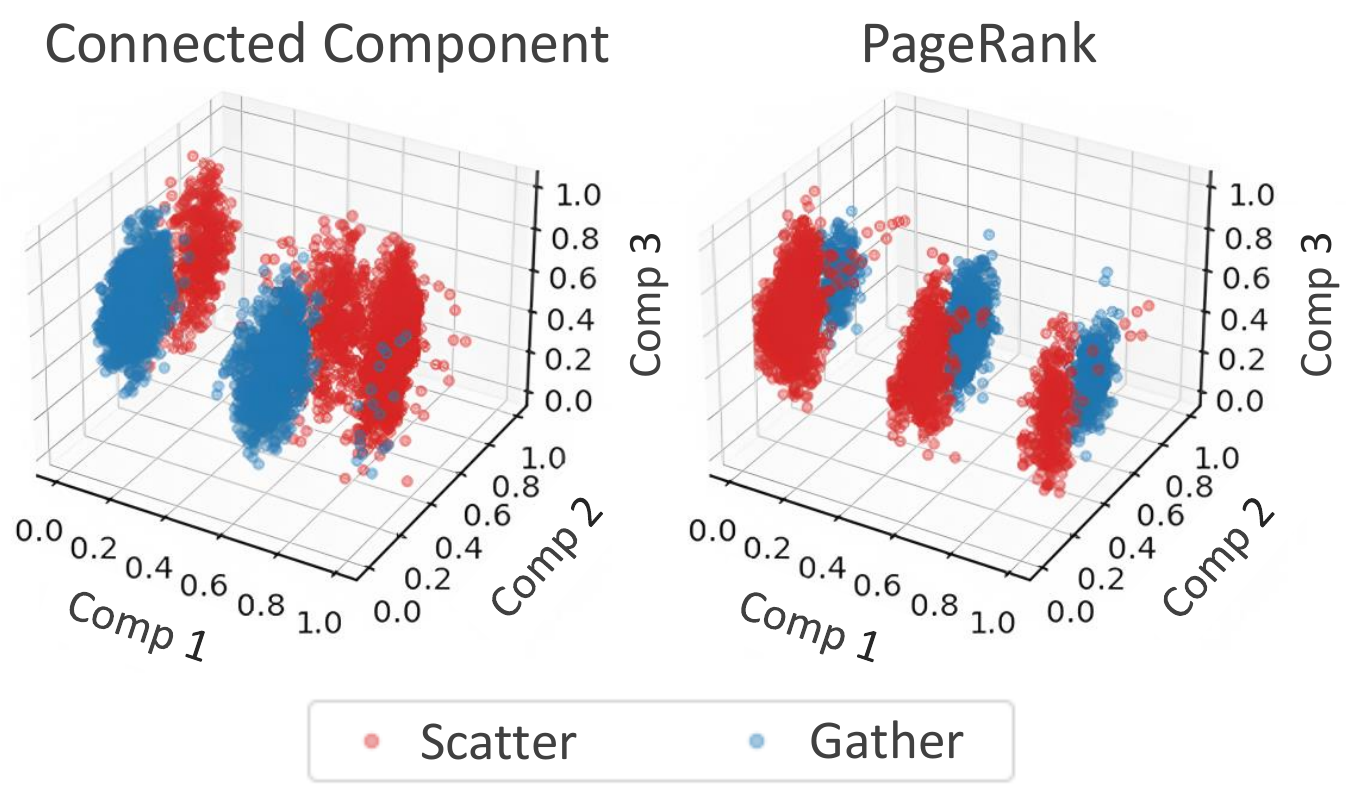}}
  \subfloat[PCA for program counters.\label{fig:pca2}]{%
        \includegraphics[width=0.5\linewidth]{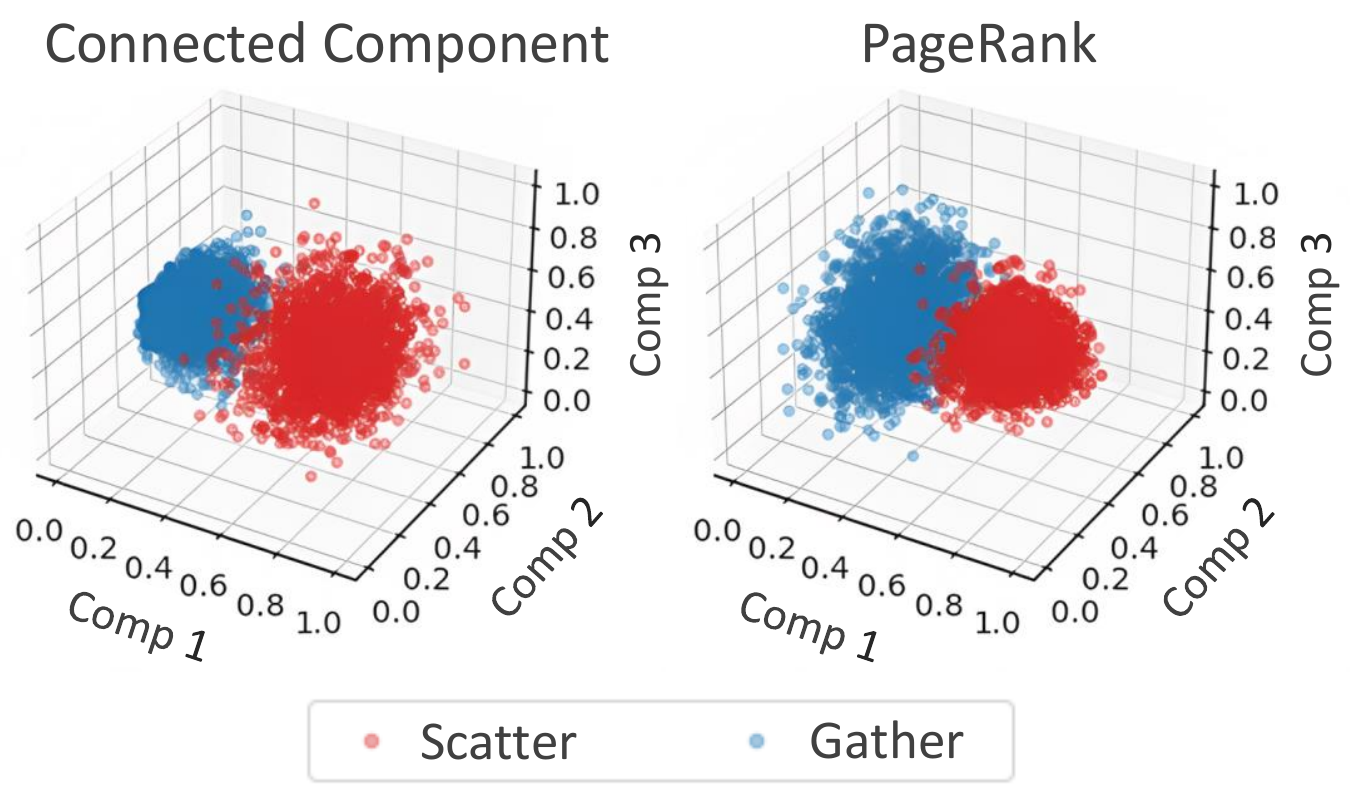}}
  \caption{Scatter and Gather phases in GPOP applications have distinct memory access patterns. The phase transitions can be detected by analyzing the program counters.}
  \label{fig:pca} 
\end{figure}

\noindent\textbf{Incorporation of phase.} Memory access patterns vary among different phases of graph processing. 
{
\color{black}
We perform Principal Component Analysis (PCA) of the memory access sequences from the applications of Connected Component and PageRank in GPOP. The results of the three top components (Comp 1-3) are shown in Figure~\ref{fig:pca1}, there is diversity in memory access patterns both within and between the two phases.
}Although the number of patterns within a phase is not fixed, the number of phases is small and constant. This observation has led us to train a separate model for each phase to improve memory access prediction performance. Additionally, Figure~\ref{fig:pca2} shows that program counters form clusters for each phase, which suggests that program counters can be used to detect phase transitions.

\noindent\textbf{Incorporation of modality.} 
Parallel execution of multi-threaded applications results in interleaved instructions and high irregularity in memory accesses. For example, GPOP processes partitions of the input graph in parallel on each core. Memory address sequence cannot fully reveal this characteristic. Instead, from an instruction perspective, a multi-threaded process has multiple PCs, each pointing to the next instruction to execute for a given thread. Therefore, we use the PC sequence as an equally important input modality as the address sequence and develop a multi-modality network to fuse the two inputs for memory access prediction.

\begin{figure} [t]
    \centering
  \subfloat[Scatter phase.\label{fig:pj1}]{%
        \includegraphics[width=0.5\linewidth]{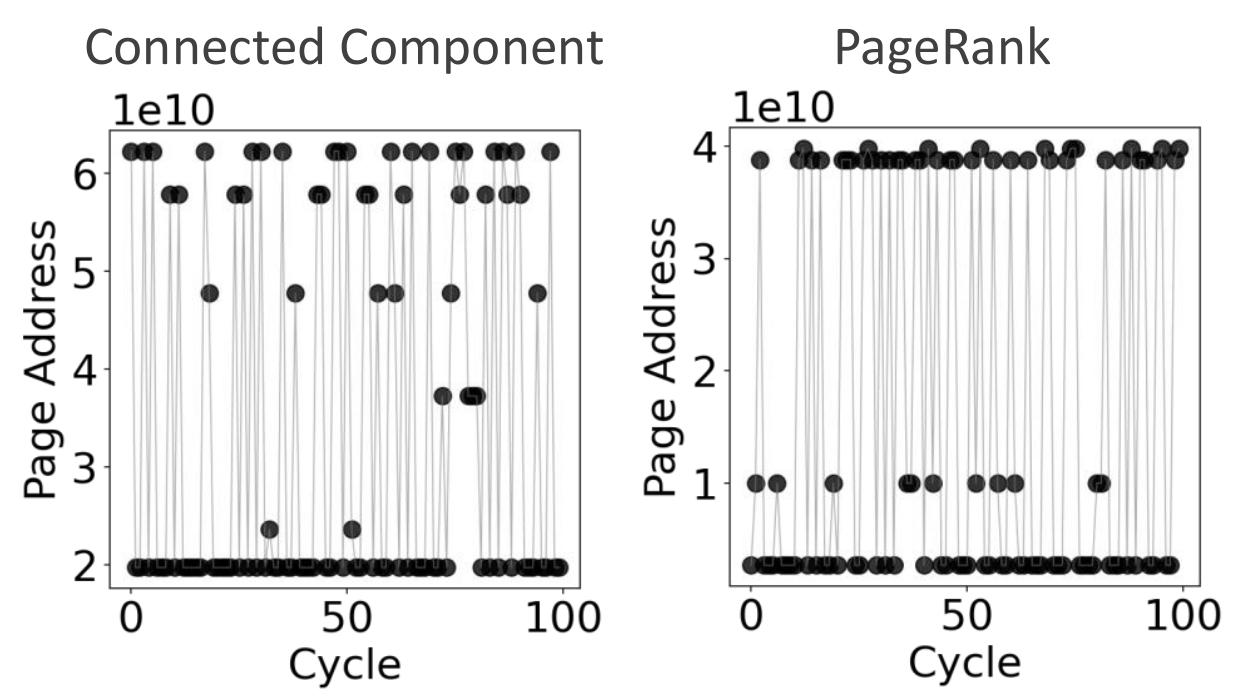}}
  \subfloat[Gather phase.\label{fig:pj2}]{%
  \includegraphics[width=0.5\linewidth]{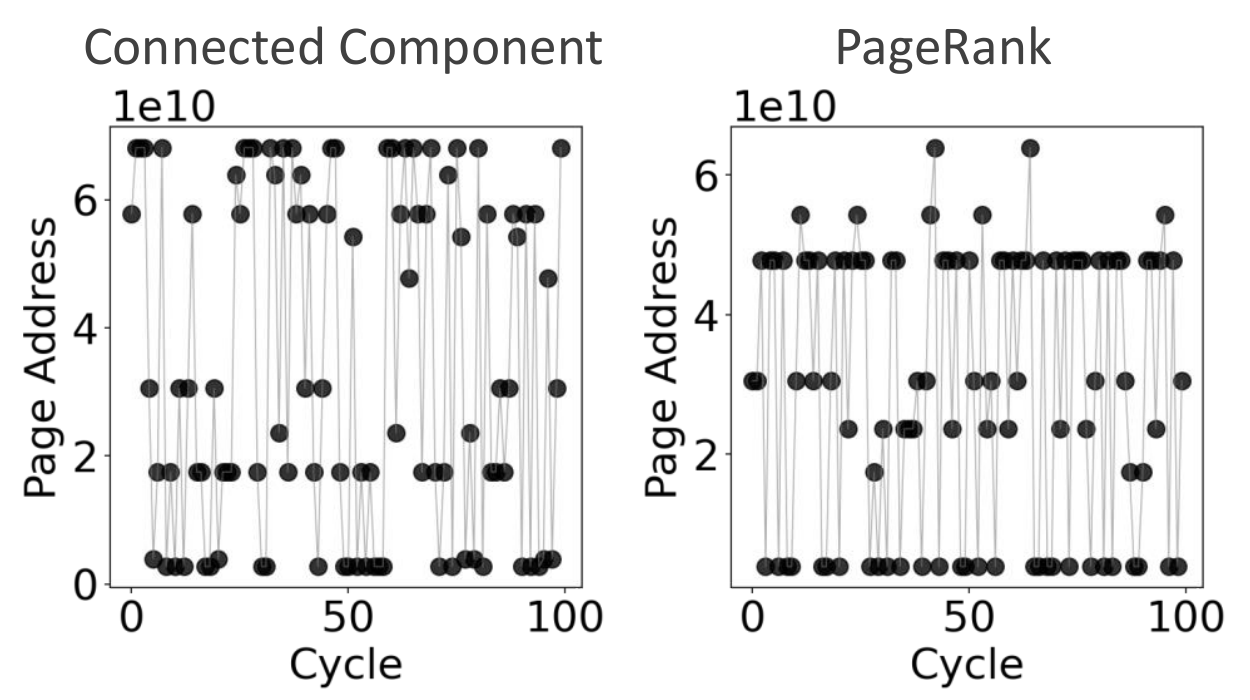}}
  \caption{Wide-range memory access page jumps in GPOP.}
  \label{fig:page_jump} 
\end{figure}

\noindent\textbf{Incorporation of locality.} Large-scale graph processing often involves high irregularity in memory access when accessing graph nodes stored in different memory pages. As shown in Figure~\ref{fig:page_jump}, there are frequent and wide memory access page jumps in the case applications in GPOP. Therefore, in addition to predict deltas within a page following spatial locality, we propose to also predict memory access pages.
Considering that graph processing is iterative, we predict pages following temporal locality.

In summary, to build a high-performance prefetcher for graph analytics, we design domain specific ML models tailored for multi-core shared-memory platforms running graph analytics applications with iterative barrier-synchronized phases. These models include a phase transition detector and phase-specific multi-modality predictors for memory access delta and page prediction. The detailed design is presented in Section~\ref{sec:approach}. 
{\color{black}
The proposed approach can be extended to accelerate 1) graph neural networks with well-defined phases (e.g. aggregation and update), 2) GPU applications that uses shared memory, and 3) HPC applications where graphs are extremely large, etc.
}

\section{Approach}
\label{sec:approach}

\subsection{Overview}
\label{sec:overview}
\begin{figure}[ht]
  \centering
  \includegraphics[width=0.98\linewidth]{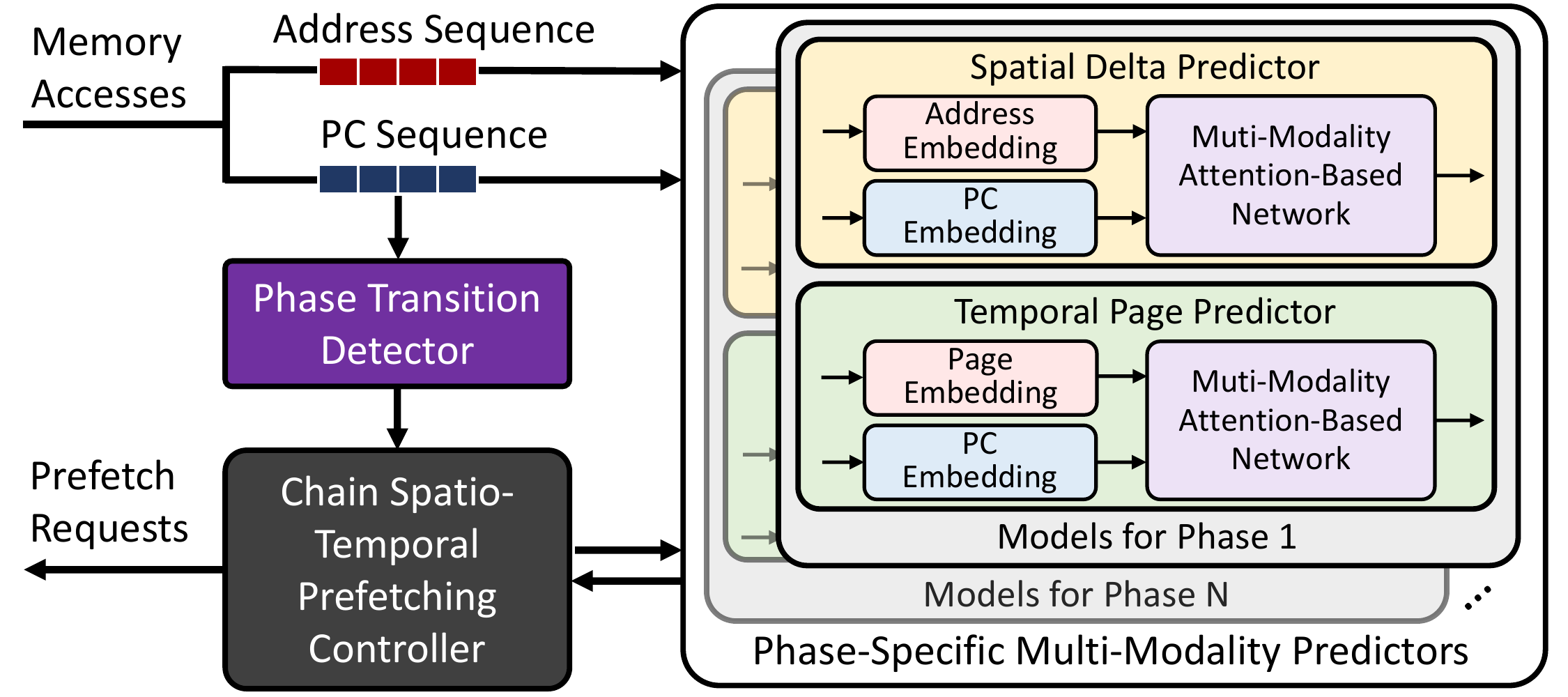}
  \caption{Overall design of~\ourwork. Domain specific ML models operate in a phase transition detector and phase-specific multi-modality predictors. A prefetching controller performs a chain spatio-temporal prefetching strategy to manage the predictors and request prefetches.
  }
  \label{fig:overall}
\end{figure}

We introduce ~\ourwork, an ML-based prefetcher designed to accelerate graph analytics frameworks using domain specific ML models. Figure~\ref{fig:overall} illustrates the overall design and workflow of~\ourwork.  
The prefetching process begins with a phase transition detector that reads the PC sequence and detects phase transitions (see Section~\ref{sec:pd}). For each phase, there are phase-specific multi-modality predictors for spatial delta prediction and temporal page prediction, both using address sequence and PC sequence as two modalities of inputs (see Section~\ref{sec:model}).
To select the phase-specific predictors and manage the delta and page predictions, we develop a prefetching controller that operates a novel chain spatio-temporal prefetching strategy and generates prefetch requests (see Section~\ref{sec:cstp}). 

\noindent\textbf{Problem Formulation.}
Given a graph analytics application that uses a paradigm with $N$ phases, we have sequences of $T$ history physical block addresses $A_t = \{a_{1}, a_{2}, ..., a_T\}$  and history program counters $PC_t = \{pc_{1}, pc_{2}, ..., pc_T\}$ at time $t$. The input for the prediction models is $X_t = \{(a_1, pc_1), (a_2, pc_2) ..., (a_T, pc_T)\}$ and the future $K$ block addresses are represented by $Y_t=\{y_1, y_2, ..., y_K\}$. Our goal is to train a phase transition detector and $N$ sets of models for each phase to learn the mapping from $X_t$ to $Y_t$. These models will then be used to perform memory access prediction and prefetching to improve instructions per cycle (IPC).

\subsection{Phase Transition Detector}
\label{sec:pd}

Phases are defined by the software and extracting phase labels requires access to the source code and software interface. With this in mind, we have developed phase transition detectors for two scenarios based on the accessibility of phase labels.

\subsubsection{Phase label Inaccessible}
\label{sec:kswin}

In scenarios where phase labels are not accessible, we use an unsupervised learning model to detect phase transitions. Given a stream of program counters at time $t$, represented as $PC_t = {pc_{1}, pc_{2}, …, pc_T}$, and the corresponding phase as class $n_t\in{0,1,…,N-1}$, the distribution of $PC_t$ is defined as $P(PC_t, n_t)$. If a transition of phase occurs at time $t$, then:
\begin{equation}
P(PC_t,n_t) \neq P(PC_{t-1},n_{t-1}) 
\end{equation}
The phase transition is described as a concept drift~\cite{lu2018learning} in the context of machine learning.

\noindent\textbf{Kolmogorov–Smirnov Windowing (KSWIN)}~\cite{raab2020reactive} is a state-of-the-art concept drift detection model for a data stream. It is based on the two sample Kolmogorov–Smirnov (K-S) test~\cite{berger2014kolmogorov}, which estimates the probability that two sets of samples were drawn from the same distribution.
The K-S statistic $D$ is the absolute distance between the two empirical cumulative distributions $F_A$ and $F_B$:
\begin{equation}
D_{A,B} = \underset{x}{\operatorname{sup}}(|F_A(x) - F_B(x)|)
\end{equation}
where $F_{(\cdot)}(x)=\frac{1}{n} \sum_{i=1}^n I_{[-\infty, x]}\left(X_i\right)$ is the empirical distribution function for $n$ ordered observations $X_i$; $I_{[-\infty, x]}\left(X_i\right)$ is an indicator function that equals to 1 if $X_i < x$ and 0 otherwise; $\text{sup}(x)$ is the supremum of the set of distances.
KSWIN detects concept drifts by comparing the distributions of two windows: one containing $h$ history samples $H$ and the other containing $r$ recent samples $R$. These windows are sampled from a sliding window $\Psi$, which keeps $w$ most recent points from the stream, as shown below:
\begin{equation}
R=\left\{\mathbf{x}_i \in \Psi\right\}_{i=w-r+1}^w
\end{equation}
\begin{equation}
H=\left\{\mathbf{x}_i \in \Psi \mid i\leq w-r; p(\mathbf{x})=\mathcal{U}\left(\mathbf{x}_i \mid 1, w-r\right)\right\}
\end{equation}

We can reject the null hypothesis (that there is no statistically significant difference between the two observations) at a significance level of $\alpha$ if the following inequality is satisfied:
\begin{equation}
    D_{H, R}>\sqrt{-\ln\left(\frac{\alpha}{2}\right) \frac{1+\frac{r}{h}}{2 r}} \overset{(h=r)}{=} \sqrt{-\ln\left(\frac{\alpha}{2}\right) \frac{1}{r}}
\end{equation}

\begin{figure}[t]
    \centering
  \subfloat[False positive in KSWIN.\label{fig:kswin1}]{%
  \raisebox{1ex}%
  {\includegraphics[width=0.36\linewidth]{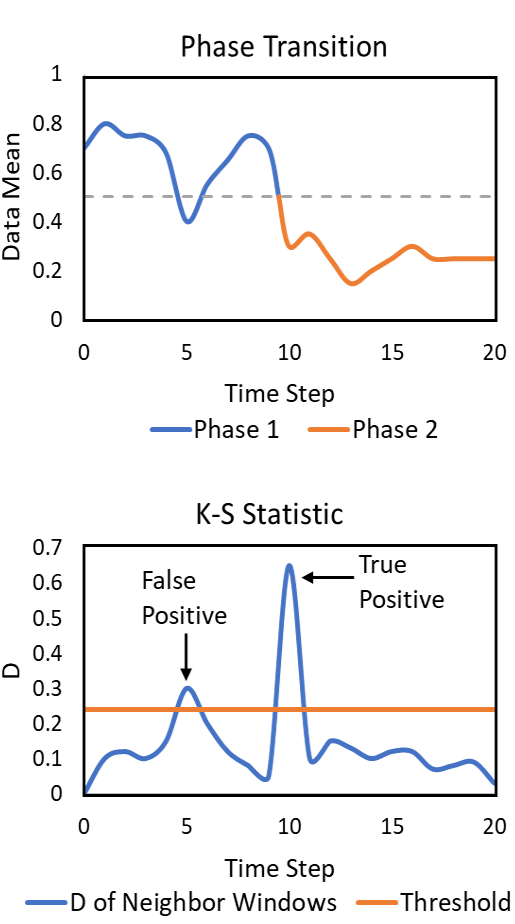}} 
  }
    \quad
  \subfloat[Sampling in KSWIN and Soft-KSWIN.\label{fig:kswin2}]{%
        \includegraphics[width=0.56\linewidth]{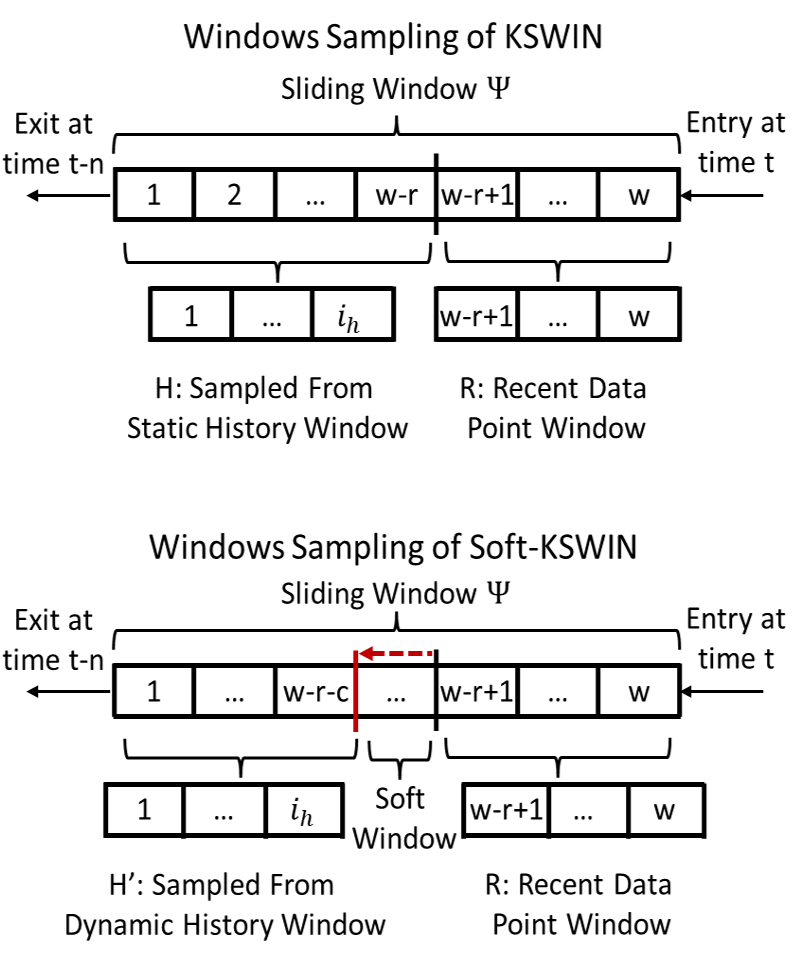}}
  \caption{
  KSWIN’s false positive issue arises from its hard detection of statistic $D$ against a threshold. Soft-KSWIN addresses this using a modified history window sampling.}

  \label{fig:kswin} 
\end{figure}

The hyperparameter $\alpha$ determines the threshold for drift detection in the KSWIN model and its selection is crucial. The model is very sensitive to the selection of $\alpha$. A large value of $\alpha$ may increase the rate of false positive detections, while a small value may cause the model to fail in detecting drifts. KSWIN reports a positive phase transition when $D$ exceeds the threshold. This “hard” detection process ignores pulsing pattern shifts within a phase and can result in false positive detections, as shown in Figure~\ref{fig:kswin1}.

\begin{algorithm}
  \caption{Soft-KSWIN for Phase Transition Detection}
  \label{alg:soft-kswin}
  \begin{algorithmic}[1]
    \State Initialize $\alpha$, $th\_r$
    \State Initialize sliding window $\Psi$ and its length limit $w$
    \State Initialize soft history window $H'$ with size $h$
    \State Initialize recent window $R$ with size $r$
    \State Initialize $counter\leftarrow0$, $detection\leftarrow 0$
    \State Initialize $transition \leftarrow False$
    \State Initialize $threshold \leftarrow \sqrt{-\ln\left(\frac{\alpha}{2}\right) \frac{1+\frac{r}{h}}{2 r}}$
    \If {length($\Psi$) $\geq w$} 
        \State Pop the oldest data point from $\Psi$
        \State Push the new data point to $\Psi$
        \State $H'\leftarrow$ Sample uniformly from $\Psi$[1:w-r-counter]
        \State $D_{H',R} \leftarrow \operatorname{sup}(|F_{H'}(x) - F_R(x)|)$ \Comment{Inference}
        \If{$D_{H',R}>threshold$} \Comment{Soft detection}
            \State $counter \leftarrow counter + 1$
            \State $detection \leftarrow detection +1$
            \If{$counter\geq r$}
               \If{$detection/counter > th\_r$}
                \State $transition \leftarrow True$
                \State Reset the model
            \EndIf
            \EndIf
        \EndIf
        \State $counter \leftarrow counter +1$
    \Else
        \State Push new data point to $\Psi$
    \EndIf
  \end{algorithmic} 
\end{algorithm}

\noindent\textbf{Soft-KSWIN} is a domain specific variant of KSWIN designed to detect graph processing phase transitions. It exploits domain knowledge that phases are stable for millions of instructions to avoid false positives. We design a soft detection process using a soft history window $H'$ sampled from a dynamic set of history data points, as shown in Figure~\ref{fig:kswin2} and Equation~\ref{eq:soft_h}:
\begin{equation}
H'=\left\{\mathbf{x}_i \in \Psi \mid i\leq w-r-c; p(\mathbf{x})=\mathcal{U}\left(\mathbf{x}_i \mid 1, w-r-c\right)\right\}
\label{eq:soft_h}
\end{equation}
where $c$ is a counter initiated when a positive detection occurs. 

The Soft-KSWIN algorithm is shown in Algorithm~\ref{alg:soft-kswin}. The history window samples from the sequence not polluted by the detected new pattern. As data flows through the stream, if phase transition detection occurs between the recent window with the old pattern, both the counter $c$ and the number of detections are incremented (lines 13-15 in Algorithm~\ref{alg:soft-kswin}). When an entirely new recent window is sampled and $counter\geq r$, if the ratio between the number of detections and the counter is larger than a soft threshold $th\_r$ (default set as 0.5), a final detection is declared to be positive and the model is reset for future detection (lines 16-20 in Algorithm~\ref{alg:soft-kswin}).

\subsubsection{Phase Label Accessible}
 
Processing phases can be labeled offline using source code, programmer annotation, and instrumentation tools like Intel Pin~\cite{luk2005pin}. A supervised model can then be trained using the PC trace and phase labels.

\noindent\textbf{Decision Tree (DT).} We use a simple decision tree classifier~\cite{myles2004introduction} to predict the current phase from the PC trace sequence. When two consecutive predictions differ, we detect a phase transition.

\noindent\textbf{Soft Decision Tree (Soft-DT).} We observe that DT can produce false positive detections when it immediately reports phase transitions upon predicting a different phase. This includes short-term pattern shifts within a phase and false predictions. To reduce the false positive rate, we use a soft detection method similar to Section~\ref{sec:kswin}. We store past phase inferences in a result queue $Q$ and compare the modes (most frequently occurred element in a list) between its head and tail halves. When the two modes differ, we report a transition detection to avoid impulse pattern shifts.

\subsection{Phase-Specific Multi-Modality Predictors}
\label{sec:model}

\begin{figure}[t]
    \centering
  \subfloat[Spatial delta predictor.\label{fig:mm-s}]{%
  \includegraphics[width=0.45\linewidth]{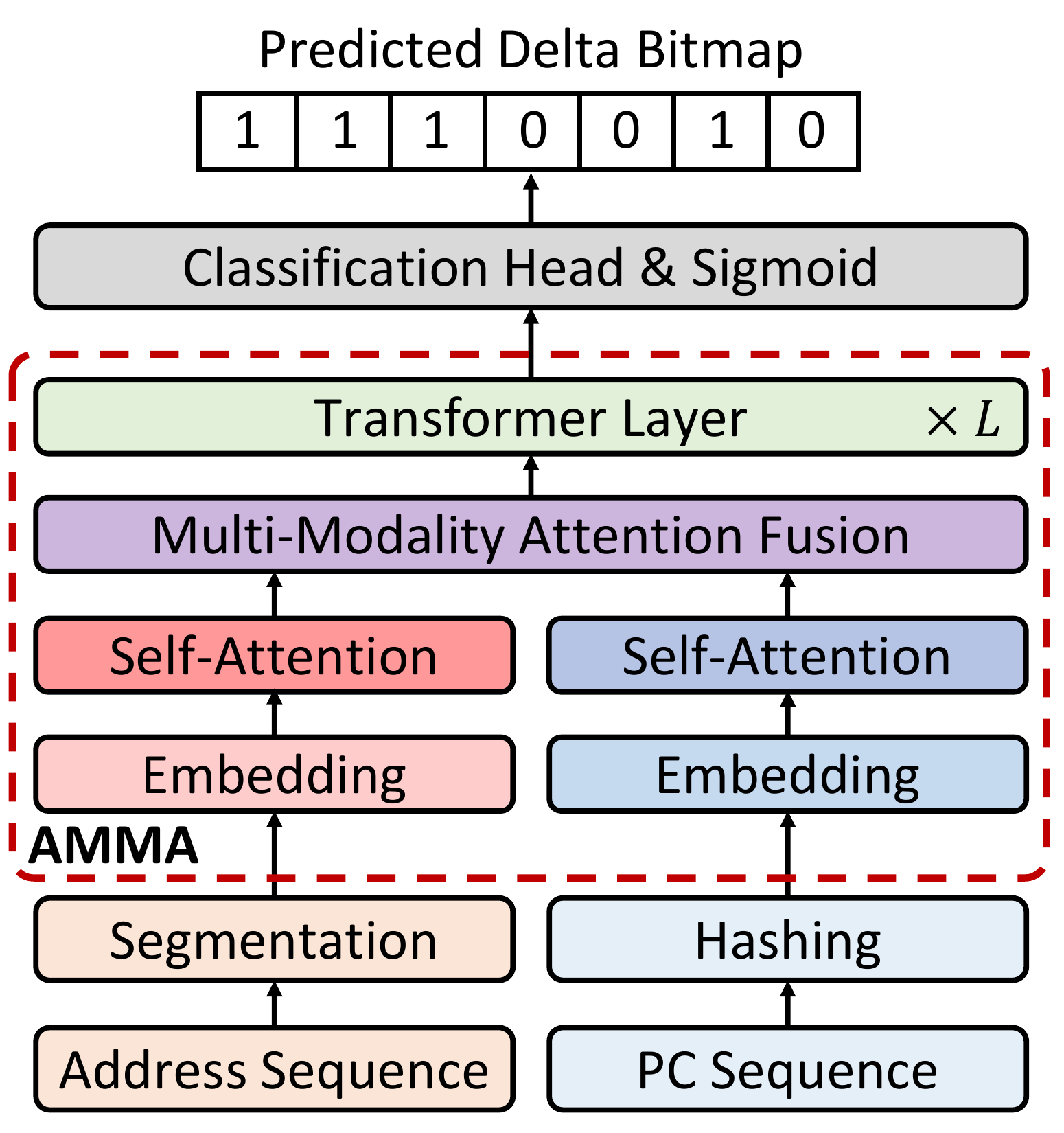}
  }
    \quad
  \subfloat[Temporal page predictor.\label{fig:mm-t}]{%
        \includegraphics[width=0.45\linewidth]{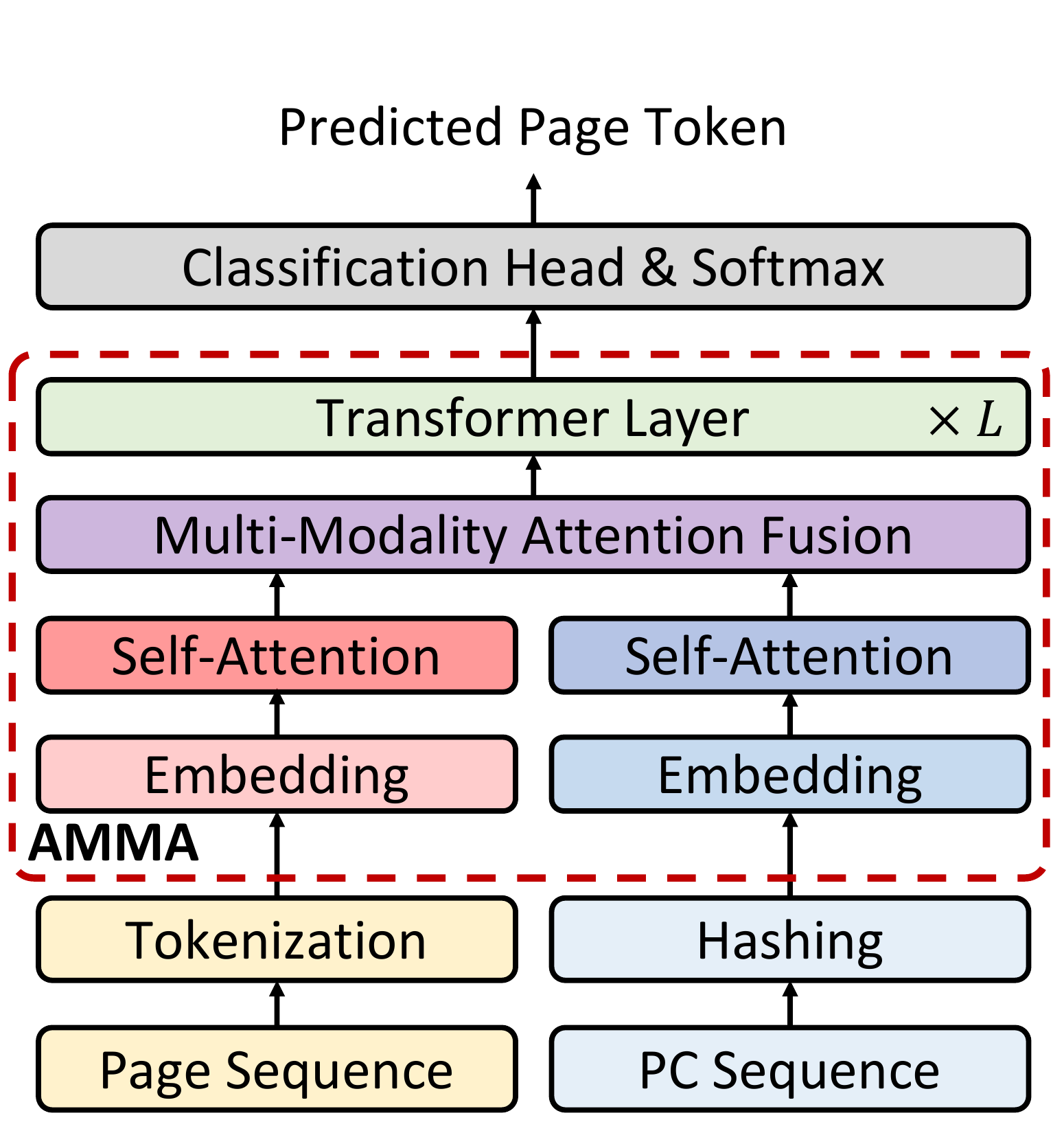}}
  \caption{Phase-specific attention-based multi-modality models for memory access delta and page predictions.}
  \label{fig:mm} 
\end{figure}

We train phase-specific predictors for each phase of graph processing. To utilize spatial and temporal locality, we design two domain specific models: the spatial delta predictor and the temporal page predictor. We develop AMMA, an \underline{A}ttention-based network with \underline{M}ulti-\underline{M}odal \underline{A}ttention fusion, as the backbone (feature extractor) for both predictors, see Figure~\ref{fig:mm}. We use attention mechanism in our models due to its high adaptability in prediction and high parallelizability in implementation~\cite{vaswani2017attention}.

\begin{figure}[h]
  \centering
  \includegraphics[width=1\linewidth]{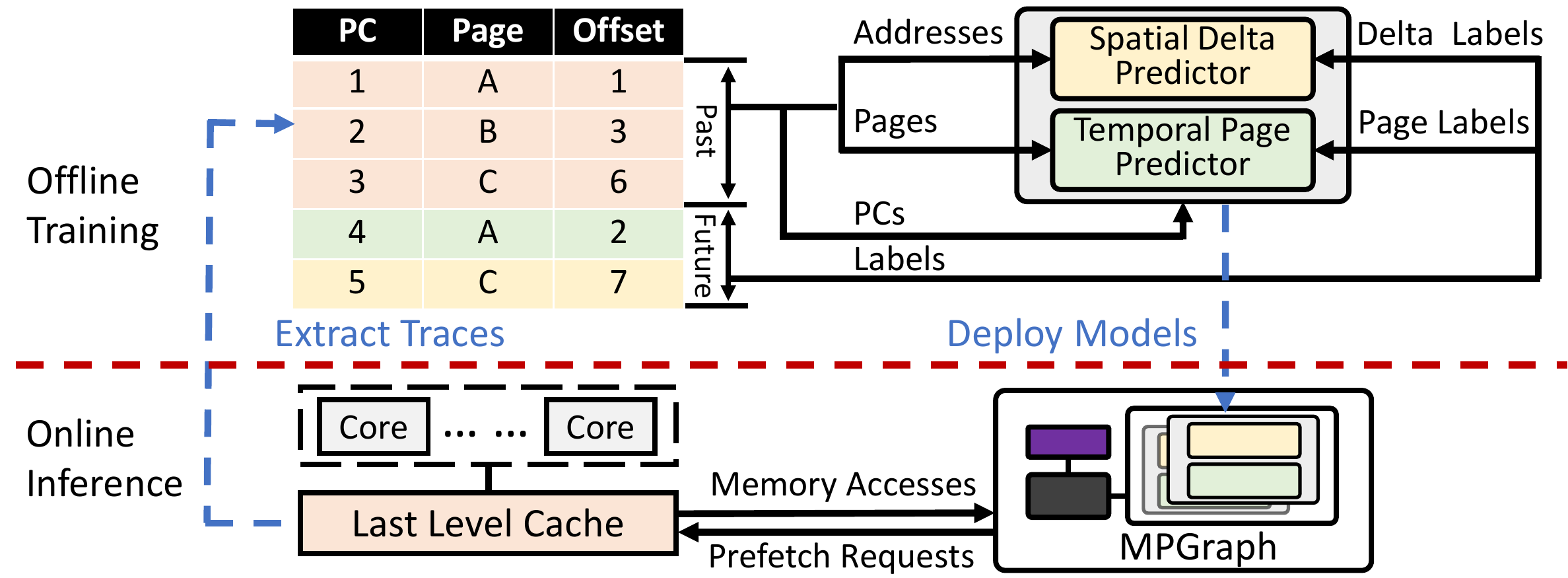}
  \caption{{\color{black}Workflow of training and inference of predictors.} }
  \label{fig:train}
\end{figure}

{\color{black}
\subsubsection{Workflow of Training and Inference of Predictors.}
The predictors in~\ourwork~are trained offline and then deployed for online inference. The workflow is shown in Figure~\ref{fig:train}. We extract memory access traces by monitoring the application access to the shared last level cache. Considering the iterative characteristic of graph analytics applications, we extract the memory access traces for the first iteration of execution. Using this extracted trace, we set scanning windows for $T$ past accesses (as input) and $F$ future accesses (to collect labels for future page and deltas) at a specific time $t$. We train phase-specific models offline using memory access traces extracted from the phases. Once the models are trained, we deploy the models to the proposed ML-based prefetcher~\ourwork~for accelerating future execution of the application.
}

\subsubsection{AMMA} 

AMMA is composed of three types of attention-based layers: a self-attention layer, a multi-modality attention fusion layer, and $L$ Transformer layers. 

\noindent{\textbf{Self-Attention Layer}} takes the embedding of items as input, converts them to three matrices through linear projection, then feeds them into a scaled dot-product attention defined as:
\begin{equation}
\label{eq:att}
\operatorname{Attention}(Q, K, V)=\operatorname{softmax}\left(\frac{Q K^{T}}{\sqrt{d_{k}}}\right) V
\end{equation}
where $Q$ represents the queries, $K$ the keys, $V$ the values, $d$ the dimension of layer input. 

\noindent{\textbf{Multi-Modality Attention Fusion Layer}} merges the input of two modalities through concatenation and self-attention:
\begin{equation}
\begin{aligned}
\label{eq:att_fusion}
E_{cx} &= \operatorname{Concat}\left(E_{1}, \ldots, E_{M}\right)\\
\operatorname{MMAF}(E_{cx}) &= \operatorname {Attention}\left(E_{cx} W_{i}^{Q}, E_{cx} W_{i}^{K}, E_{cx} W_{i}^{V}\right)
\end{aligned}
\end{equation}
{\color{black}
where $E$ is the input of the MMAF layer, $W$ is the weight matrix for a fully connected layer, $M$ is the number of modalities.} $M=2$ in this work, one modality is the address sequence and the other is the PC sequence.

\noindent\textbf{Transformer Layer} consists of a Multi-head Self Attention ($\operatorname{MSA}$) and a point-wise feed-forward network ($\operatorname{FFN}$). 
\begin{equation}
\begin{aligned}
\label{eq:msa}
\operatorname{MSA}(Q, K, V) &=\operatorname{Concat}\left(\operatorname{head}_{1}, \ldots, \text {head}_{\mathrm{H}}\right) W^{O} \\
\text{head}_{\mathrm{i}} &=\text {Attention}\left(Q W_{i}^{Q}, K W_{i}^{K}, V W_{i}^{V}\right)
\end{aligned}
\end{equation}
\begin{equation}
\label{eq:att3}
\operatorname{FFN}(x)=\max \left(0, x W_{1}+b_{1}\right) W_{2}+b_{2}
\end{equation}
where $W_{i}^{Q}, W_{i}^{K}, W_{i}^{V} \in \mathbb{R}^{d \times d}$ and H is the number of heads. 

\subsubsection{Spatial Delta Predictor}
Figure~\ref{fig:mm-s} illustrates the model of the spatial delta predictor. It uses both the address sequence and PC sequence as inputs, with AMMA serving as the backbone and a Multi-Layer Perceptron  (MLP) 
with a Sigmoid function acting as the classification head.

\noindent\textbf{Input.} The input preprocessing employs the address segmentation method~\cite{zhang2022fine}. An input memory address is divided into a list of segments to make it processable by an ML model and to avoid tokenizing the vast space of addresses (in millions). The PC is hashed and normalized for processing by the model.

\noindent\textbf{Output.} The model predicts multiple future deltas within a spatial range, i.e., a page size. The sum of the current address and the predicted delta is the prediction of future accesses.

\noindent\textbf{Training.} The model is trained using labels from future deltas in the form of a bitmap for multi-label classification. We use binary cross entropy as the loss function.

\subsubsection{Temporal Page Predictor}
Figure~\ref{fig:mm-t} illustrates the model of the temporal page predictor. It takes the page address part of a memory address as a modality and uses PC as another modality. AMMA serves as the feature extractor and an MLP with a Softmax activation function is used for the classification head.

\noindent\textbf{Input.} The page sequence input is tokenized because the small vocabulary (in thousands) can be processed by an ML model~\cite{shi2021hierarchical}. 

\noindent\textbf{Output.} The model outputs the probability of the next future page as determined by the Softmax function.

\noindent\textbf{Training.} The model is trained using the future page token as label. We use categorical cross entropy as the loss function.

\subsection{Chain Spatio-Temporal Prefetching Controller}
\label{sec:cstp}
The prefetching controller performs two functions: it switches between phase-specific predictors and specifies the prefetch request. We propose a novel strategy called Chain Spatio-Temporal Prefetching (CSTP) to determine what to prefetch.
\subsubsection{Switching Predictors}
\label{sec:switch}
The prefetching controller receives signals from the phase transition detector. Upon detecting a transition, it activates all $N$ phase-specific predictors to work in parallel. The controller then monitors the performance of these predictors for a small number of accesses and selects the best-performing one.

\subsubsection{Chain Spatio-Temporal Prefetching}
\label{sec:chain}
\begin{figure}[h]
  \centering
  \includegraphics[width=1\linewidth]{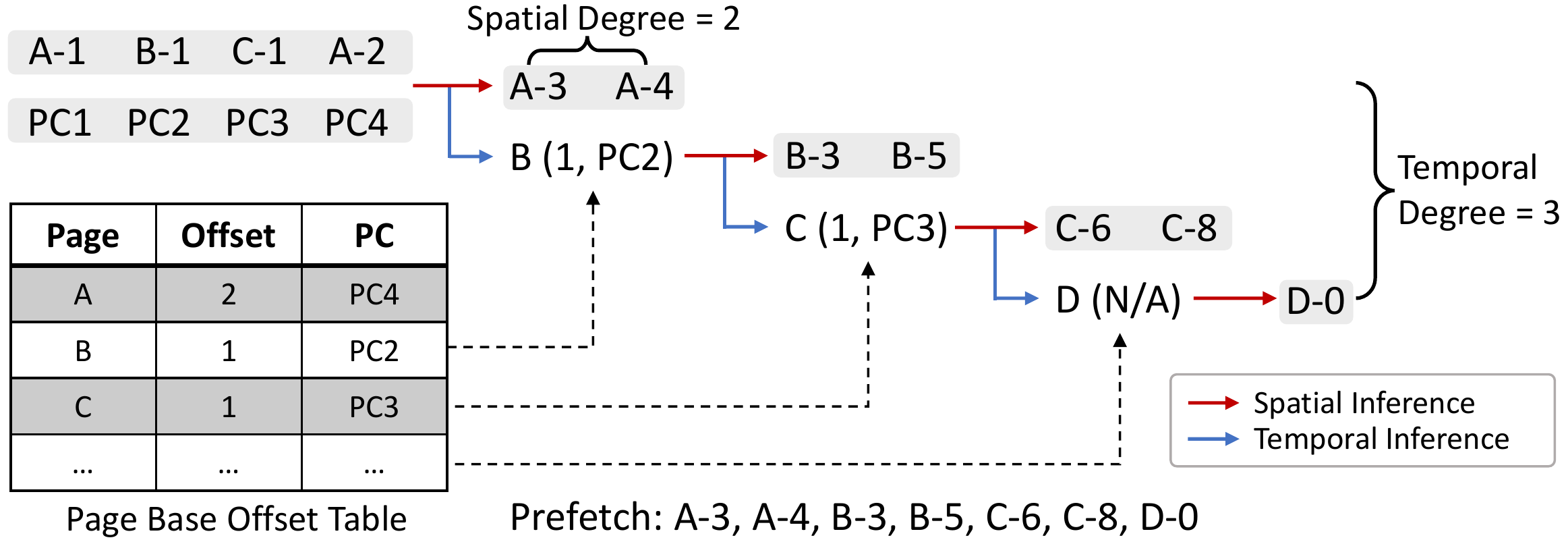}
  \caption{Chain spatio-temporal prefetching. }
  \label{fig:chain}
\end{figure}

Figure~\ref{fig:chain} illustrates the CSTP strategy. Given an input sequence of memory block addresses denoted using "Page-offset" (e.g., A-1 represents page A offset 1) and a PC sequence, the spatial delta predictor and temporal page predictor operate in parallel. A page base offset table (PBOT) records the latest offset and PC for past pages. For a predicted page, the latest offset and PC can be retrieved from the PBOT for further spatial and temporal inference. This process continues as a chain until either the temporal degree is exceeded or the page offset is missing in PBOT. Given a spatial degree $D_s$ and temporal degree $D_t$, the total prefetch degree $D_p$ range is:
\begin{equation}
\label{eq:degree}
D_s + 1\leq D_p\leq D_s(D_t+1)
\end{equation}

\begin{figure*}[h]
  \centering
  \includegraphics[width=1\linewidth]{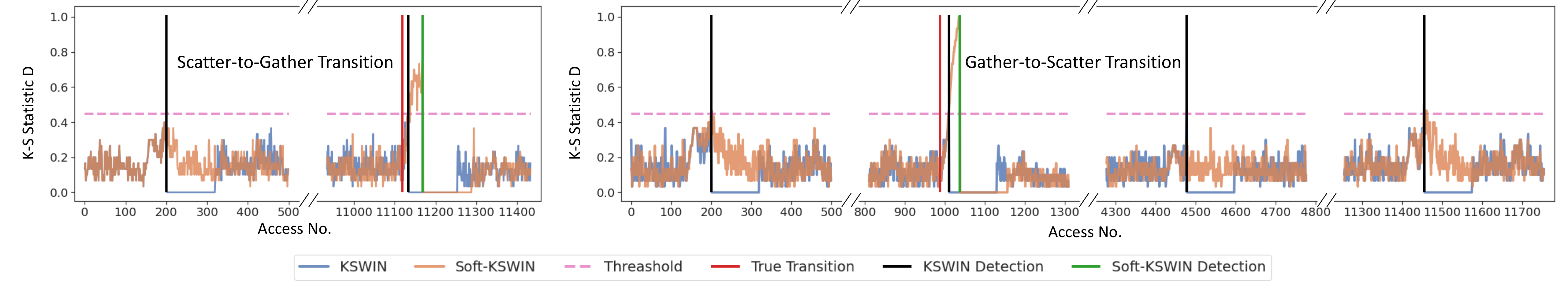}
  \caption{Case study on the phase detection performance of KSWIN and Soft-KSWIN on GPOP PageRank. KSWIN reports false positive results due to "hard" detection. Soft-KSWIN avoid the false positives while incurring a small window of lag.}
  \label{fig:ks-case}
\end{figure*}

 \section{Evaluation}

\subsection{Experimental Setup}

\subsubsection{Benchmarks}
We evaluate~\ourwork~and the baselines using three popular graph processing frameworks: GPOP~\cite{lakhotia2020gpop}, X-Stream~\cite{xstream}, and PowerGraph~\cite{gonzalez2012powergraph}. We use the built-in applications of the frameworks for evaluation, including Breadth-First Search~(BFS), Connected Components~(CC), PageRank~(PR), Single Source Shortest Path~(SSSP), and Triangle Counting~(TC). Details of are shown in Table~\ref{tab:benchmark}.

\begin{table}[htbp]
	\centering
	\caption{Benchmark Graph Frameworks and Applications}
	\label{tab:benchmark}
		\begin{tabular}{|c|c|c|c|}
		\hline
		\textbf{Framework} & \textbf{Paradigm} & \textbf{$N^\ast$} &\textbf{Applications} \\ \hline
		GPOP~\cite{lakhotia2020gpop} & Scatter-Gather & 2 & BFS, CC, PR, SSSP\\\hline
		X-Stream~\cite{xstream} & Scatter-Gather & 2 & BFS, CC, PR, SSSP\\\hline
		PowerGraph~\cite{gonzalez2012powergraph} & GAS & 3 & CC, PR, SSSP, TC\\
		\hline
		\multicolumn{4}{l}{\small $\ast$ $N$: number of phases per iteration.}\\
		\end{tabular}
\end{table}
\subsubsection{Datasets}
We evaluate~\ourwork~and the baselines using 6 real world graph datasets~\cite{snapnets} and a randomly generated synthetic graph based on R-MAT~\cite{chakrabarti2004r}, as summarized in Table~\ref{table:datasets}.
\begin{table}[htbp]
	\centering
	\caption{Graph Datasets}
	\label{table:datasets}
		\begin{tabular}{|c|c|c|c|}
			\hline
			\textbf{Dataset} & \textbf{Description} & \textbf{\# Vertices} & \textbf{\# Edges} \\ \hline
			{amazon}~\cite{leskovec2007dynamics}     & Purchase networks     & 0.26 M                   &  1.23 M    \\ \hline
			{google}~\cite{leskovec2009community}           & Google web graph      & 0.88 M                   & 5.11 M \\ \hline
			{roadCA}~\cite{leskovec2009community}           & California Road      & 1.96 M                   &  2.76 M    \\ \hline
			{soclj}~\cite{leskovec2009community}           & LiveJournal      & 4.84 M                   &  68.99 M    \\ \hline
			{wiki}~\cite{yin2017local}           & Wikipedia hyperlinks     & 1.79 M                   &  28.51 M    \\ \hline
			{youtube}~\cite{yang2015defining}      & Youtube network    & 1.13 M                   & 2.99 M \\ \hline
		    {rmat}~\cite{chakrabarti2004r}       & synthetic graph & 1M         & 16M \\ \hline
		\end{tabular}
\end{table}
\subsubsection{Simulator}
We use ChampSim~\cite{ChampSim} for physical address traces generation and prefetcher evaluation. 
The simulator parameters are detailed in Table~\ref{tab:sim_param}. It’s important to note that while our graph datasets can be stored in DRAM,
they do not fit in the LLC.
\begin{table}[h!]
\centering
\caption{Simulation Parameters}
  \begin{tabular}{|c|c|}
    \hline
    \textbf{Parameter}&\textbf{Value}\\
    \hline
    CPU & 4 GHz, 4 cores, 4-wide OoO, \\
      & 256-entry ROB, 64-entry LSQ\\
    \hline
    L1 I-cache & 64 KB, 4-way, 4-cycle latency\\
     \hline
    L1 D-cache & 64 KB, 4-way, 4-cycle latency\\
     \hline
    L2 Cache&	512 KB, 8-way, 10-cycle latency\\
     \hline
    LL Cache&	2 MB, 16-way, 20-cycle latency\\
     \hline
    DRAM &	 $t_{RP}=t_{RCD}=t_{CAS} = 12.5$ ns, 2 channels,\\
&  8 ranks, 8 banks, 32K rows, 8GB/s bandwidth\\
  \hline
\end{tabular}
\label{tab:sim_param}
\end{table}

\begin{table}[h]
	\centering
	\caption{Phase Detection Evaluation}
	\label{tab:pd_ev}
		\begin{tabular}{|c|c|c|c|c|c|}
		\hline
			\textbf{Framework} & \textbf{Train} & \textbf{Detector} & \textbf{P} & \textbf{R}& \textbf{F1}\\\hline
			        & U$^\dagger$ &  KSWIN      &    0.2678  & 1     & 0.4224    \\\cline{3-6} 
			{GPOP}  &  &  Soft-KSWIN      &   0.8119     & 1     & 0.8958 \\ \cline{2-6}
			        & S$^\ast$ & DT   & 0.1896   &   1 &       0.3187\\\cline{3-6} 
		            && Soft-DT  &    0.8686   & 1      & 0.9297     \\\hline	       
					
					&U & KSWIN  &        0.0324  & 1     & 0.0627  \\\cline{3-6} 
			{X-Stream}  & &Soft-KSWIN      &    0.6924   & 1    & 0.8182  \\ \cline{2-6} 
			        & S &  DT & 0.1251   & 1   &   0.2238    \\\cline{3-6} 
		            &  & Soft-DT &    0.9466    & 1      & 0.9726 \\\hline	
		            
					&  U &  KSWIN  &     0.1510   & 1      & 0.2623    \\\cline{3-6} 
			{PowerGraph}  & & Soft-KSWIN      &    0.4927   & 1   &  0.6601 \\ \cline{2-6} 
			        &S &  DT &   0.0167  & 1    &   0.0328 \\\cline{3-6} 
		            & &  Soft-DT &   0.7026  & 1      & 0.8253   \\\hline	
    \multicolumn{6}{l}{\small $\dagger$ Unsupervised learning, i.e., without labeled phases.} \\
    \multicolumn{6}{l}{\small $\ast$ Supervised learning, i.e., with labeled phases.}
		\end{tabular}
\end{table}

\begin{table}[h]
  \caption{AMMA model configuration}
  \label{tab:model}
  \begin{center}
  \begin{tabular}{|l|c|l|c|}
    \hline  
    \textbf{Configuration}&\textbf{Value}&\textbf{Configuration}&\textbf{Value}\\\hline
    History $T$ & 9 & Look-forward $F$ & 256\\\hline
    Attention dimension& 64 & Fusion dimension & 128 \\\hline
    Transformer dimension & 128 & Transformer layer & 1\\\hline
    Transformer heads & 4 & MLP head layer & 1\\\hline
\end{tabular}
\end{center}
\end{table}

\subsubsection{Trace Generation}
We use Intel Pin~\cite{luk2005pin} to extract instruction traces from the benchmark applications running on 4 core.
Then, we use ChampSim to extract the shared LLC memory access trace. For model training, we use the trace from the first iteration of computation in the frameworks. For model testing and prefetching simulation, we use the traces from the following 10 iterations.

\begin{table*}[t]
	\centering
	\caption{F1-Score of Spatial Delta Prediction}
	\label{table:delta}
		\begin{tabular}{|c|c|c|c|c|c|c|c|c|c|c|c|c|}
			\hline
			&\multicolumn{12}{c|}{\textbf{Applications}} \\
			\cline{2-13}
		\textbf{Model}	&\multicolumn{4}{c|}{\textbf{GPOP}}&\multicolumn{4}{c|}{\textbf{X-Stream}}&\multicolumn{4}{c|}{\textbf{PowerGraph}}\\
			\cline{2-13}
			&\textbf{BFS}&\textbf{CC}&\textbf{PR}&\textbf{SSSP}&\textbf{BFS}&\textbf{CC}&\textbf{PR}&\textbf{SSSP}&\textbf{CC}&\textbf{PR}&\textbf{SSSP}&\textbf{TC}\\
			\hline
			LSTM & 0.2260 & 0.4041& 0.4682 & 0.2905  & 0.4964& 0.4517& 0.4719& 0.5044& 0.2096 & 0.3258 &0.3378 & 0.3744 \\\hline
			Attention & 0.2895 & 0.5259& 0.5977& 0.3643 & 0.5894 & 0.5620& 0.5577& 0.6209& 0.2965 & 0.4037 & 0.4389 & 0.4207\\\hline
			AMMA &0.3017 & 0.5976& 0.6518 & 0.3956 & 0.6027 & 0.5737& 0.5858&  0.6521 & 0.3424& 0.4180& 0.4627& 0.4341\\\hline
			AMMA-PI &0.3023 & 0.6012& 0.6501 & 0.3988 & 0.6041 & 0.5807& 0.5809&  0.6555 & 0.3511& 0.4171& 0.4694& 0.4411\\\hline
			\textbf{AMMA-PS} &\textbf{0.3375} & \textbf{0.6442}& \textbf{0.6879} & \textbf{0.4259}& \textbf{0.6346} & \textbf{0.6303}& \textbf{0.6181}& \textbf{0.6780} & \textbf{0.3973}& \textbf{0.4491}& \textbf{0.4873}& \textbf{0.4931}\\\hline
	\end{tabular}
\end{table*}

\begin{table*}[t]
	\centering
	\caption{Accuracy@10 of Temporal Page Prediction}
	\label{table:page}
		\begin{tabular}{|c|c|c|c|c|c|c|c|c|c|c|c|c|}
			\hline
			&\multicolumn{12}{c|}{\textbf{Applications}} \\
			\cline{2-13}
		\textbf{Model}	&\multicolumn{4}{c|}{\textbf{GPOP}}&\multicolumn{4}{c|}{\textbf{X-Stream}}&\multicolumn{4}{c|}{\textbf{PowerGraph}}\\
			\cline{2-13}
			&\textbf{BFS}&\textbf{CC}&\textbf{PR}&\textbf{SSSP}&\textbf{BFS}&\textbf{CC}&\textbf{PR}&\textbf{SSSP}&\textbf{CC}&\textbf{PR}&\textbf{SSSP}&\textbf{TC}\\
			\hline
			LSTM & 0.3047 & 0.4586 & 0.3336& 0.1752 & 0.3673& 0.2142& 0.1809& 0.3607&0.4105 & 0.4250 &0.5155 &0.5159 \\\hline
			Attention &0.4286 & 0.4666& 0.3339& 0.2381 & 0.3842& 0.2323& 0.1932& 0.3831& 0.4350& 0.4481& 0.5771& 0.5902\\\hline
			AMMA &0.4481 & 0.4717& 0.3790& 0.2645& 0.3999& 0.2522& 0.2168& 0.4116& 0.4495& 0.4841& 0.5996& 0.6367\\\hline
			AMMA-PI & 0.4379 & 0.4810& 0.3823& 0.2661& 0.4123& 0.2587& 0.2250& 0.4204& 0.4486& 0.4883& 0.6019& 0.6435\\\hline
			\textbf{AMMA-PS} & \textbf{0.5095} & \textbf{0.6284}& \textbf{0.4159}& \textbf{0.2848}& \textbf{0.5236}&\textbf{0.3137}& \textbf{0.3091}& \textbf{0.5864}& \textbf{0.6610}& \textbf{0.5703}&\textbf{0.6525}& \textbf{0.6571}\\\hline

	\end{tabular}
\end{table*}

\subsection{Evaluation of Phase Transition Detection}

\subsubsection{Metrics}
We use Precision (P), Recall (R), and F1-score (F1)~\cite{powers2020evaluation} to evaluate the phase transition detectors.

\subsubsection{Results}

Table~\ref{tab:pd_ev} presents the performance of various phase transition detectors. All detectors can accurately detect all true phase transitions, resulting in a recall of 1. The challenge lies in the precision of detection: false positives are common during a phase. By avoiding false positives, Soft-KSWIN achieves up to 66\% higher precision compared to KSWIN and Soft-DT achieves up to higher 82.15\% precision compared to DT. 

Figure~\ref{fig:ks-case} shows how Soft-KSWIN avoids false positives in detail. While KSWIN quickly reports a phase detection when the K-S statistic D exceeds a threshold, it also reports multiple false positive predictions due to impulsive pattern shifts. Simply setting a higher threshold cannot solve this problem and may cause true transition detection to be missed. Soft-KSWIN uses soft detection to avoid false predictions while incurring only a small lag. Since the number of instructions in a phase is in the millions, this lag is acceptable.

\subsection{Evaluation of Multi-Modality Predictors}

\subsubsection{Baselines}
We implement AMMA using configurations as in Table~\ref{tab:model}. We implement several models to demonstrate the effectiveness of our approach, including:
\begin{itemize}
    \item \textbf{LSTM}~\cite{hochreiter1997long} that use a concatenation of address and PC input for each time step. The hidden dimension is 256.
    \item \textbf{Attention}~\cite{vaswani2017attention,zhang2022fine} that use address as input and PC as side information. It uses 2 Transformer layers with dimension 128 and head number 4.
    \item \textbf{AMMA} that uses attention layers for each modality with dimension 64, the attention fusion and Transformer layer has dimension 128, each uses 1 layer.
    \item \textbf{AMMA-PI} refers to Phase-Informed AMMA, in which phase embeddings are incorporated as side information after the fusion of the two modalities in AMMA.
    \item \textbf{AMMA-PS} refers to Phase-Specific AMMA, in which separate AMMA models are trained specifically for each phase.
\end{itemize}

\subsubsection{Metrics}
We use F1-Score to evaluate the performance of spatial delta prediction, which performs multi-label classification. We use accuracy-at-10 (accuracy@10) as in~\cite{hashemi2018learning} to evaluate page prediction: a prediction is considered correct if the predicted page occurs within the next 10 memory accesses.

\begin{figure}[t]
  \centering
  \includegraphics[width=1\linewidth]{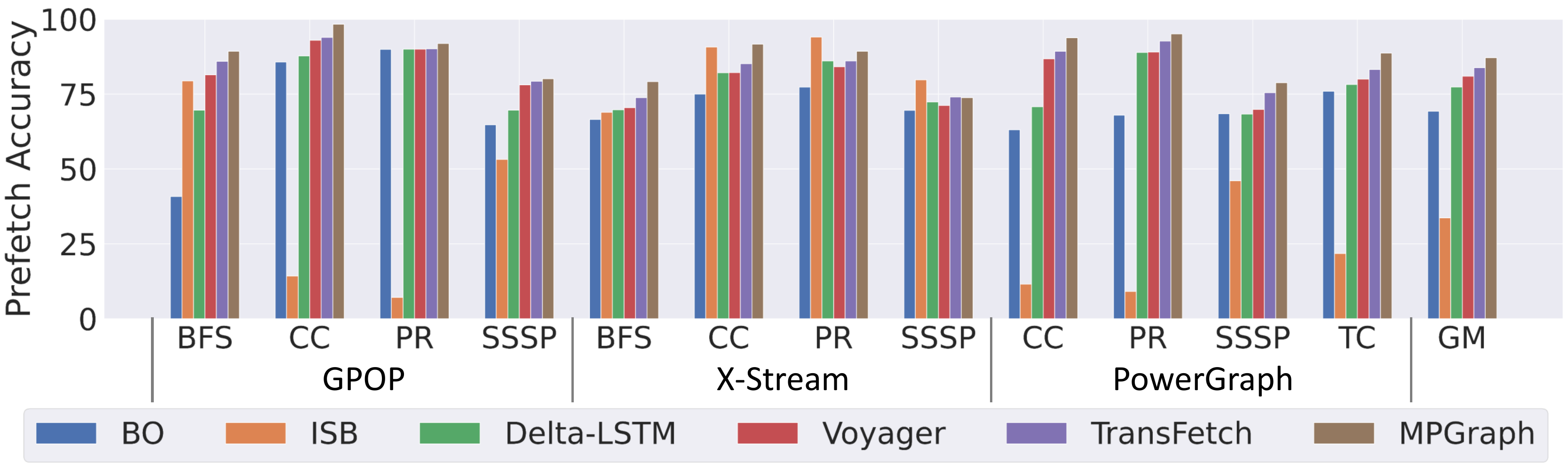}
  \caption{Prefetch accuracy of~\ourwork~and the baselines.}
  \label{fig:acc}
\end{figure}
\begin{figure}[t]
  \centering
  \includegraphics[width=1\linewidth]{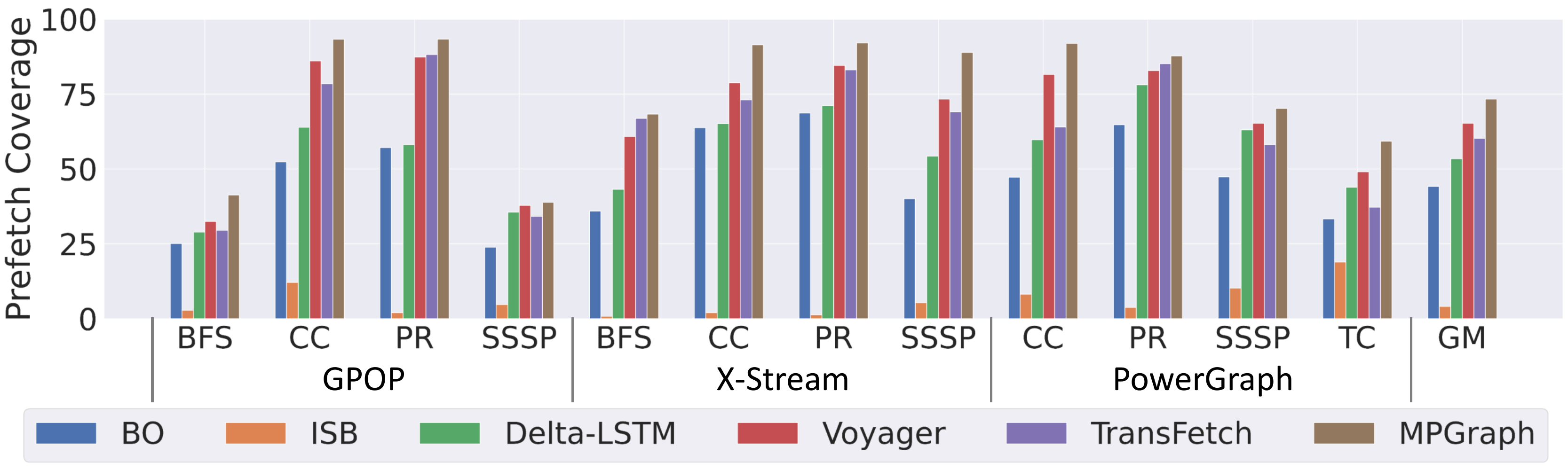}
  \caption{Prefetch coverage of~\ourwork~and the baselines.}
  \label{fig:cov}
\end{figure}

\begin{figure*}[t]
  \centering
  \includegraphics[width=1\linewidth]{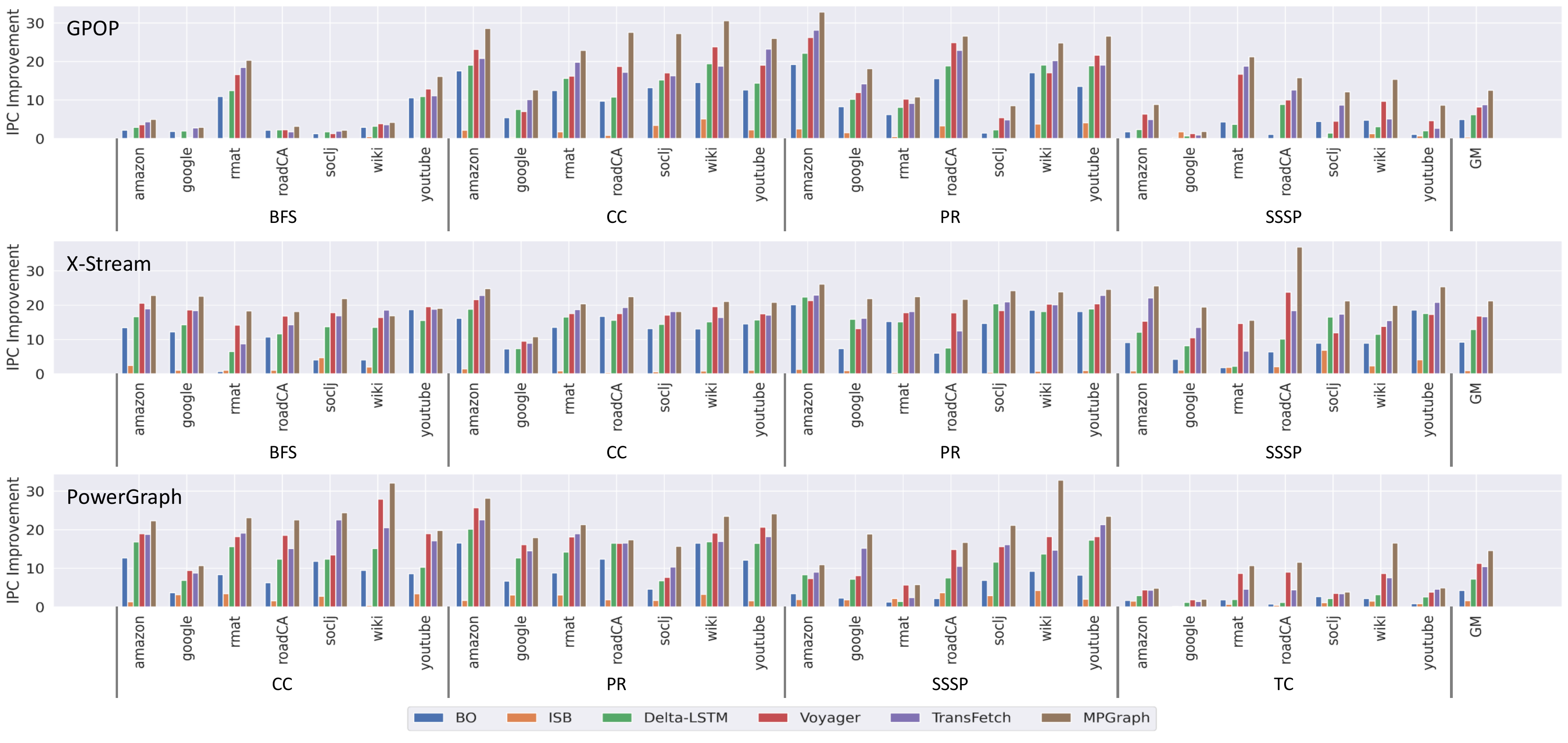}
  \caption{IPC improvement using~\ourwork~and the baselines.}
  \label{fig:ipc}
\end{figure*}

\subsubsection{Results}
Table~\ref{table:delta} shows the spatial delta prediction performance. AMMA-PS shows the highest F1-score for all the applications, outperforming LSTM by 11.15\%--24.01\%, Attention by 4.52\%--11.83\%, AMMA by 2.46\%--5.9\% and AMMA-PI by 1.79\%--5.2\%. 
Table~\ref{table:page} shows the temporal page prediction performance. AMMA-PS also shows the best performance with respect to accuracy@10 for all the applications, outperforming LSTM by 8.23\%-25.5\%, Attention by 4.67\%--22.6\%, AMMA by 2.03\%--21.15\%, and AMMA-PI by 1.36\%--21.24\%. 
Phase-specific models are particularly advantageous for page prediction due to the unique temporal patterns in each phase.

\subsection{Prefetching Evaluation}
\subsubsection{Baselines}

We compare~\ourwork~with state-of-the-art rule-based prefetchers and ML-based prefetchers:
\begin{itemize}
    \item \textbf{Best-Offset prefetcher (BO)}~\cite{michaud2016best}: rule-based spatial prefetcher that predicts delta patterns within a page.
    \item \textbf{Irregular Stream Buffer (ISB)}~\cite{jain2013linearizing}: rule-based temporal prefetcher based on record and replay.
    \item \textbf{Delta-LSTM}~\cite{hashemi2018learning}: ML-based prefetcher using delta inputs and delta outputs.
    \item \textbf{Voyager}~\cite{shi2021hierarchical}: ML-based prefetcher using address and PC as inputs and using two LSTM-based models to predict temporally the next page and offset.
    \item \textbf{TransFetch}\cite{zhang2022fine}: ML-based prefetcher using address and PC as inputs and using an attention-based model to predict deltas within a spatial range beyond a page.
\end{itemize}

We set the spatial and temporal degree of~\ourwork~to 2 and total degree 6 (Equation~\ref{eq:degree}). We set the degree of all baselines as 6.

\subsubsection{Metrics}
We use prefetch accuracy, prefetch coverage, and IPC improvement~\cite{srinivasan2004prefetch} to evaluate the prefetching performance.

\subsubsection{Results}
Figure~\ref{fig:acc} shows the prefetch accuracy averaged by input graphs for each application.~\ourwork~achieves 87.19\% accuracy, outperforming BO by 17.91\%, ISB by 53.4\%, Delta-LSTM by 9.81\%, Voyager by 6.17\%, and TransFetch by 3.35\%.    
Figure~\ref{fig:cov} shows the average prefetch coverage for each application.~\ourwork~achieves 73.29\% coverage, ourperforming BO by 29.15\%, ISB by 69.13\%, Delta-LSTM by 19.89\%, Voyager by 8.02\%, and TransFetch by 12.99\%. 
ISB shows both low accuracy and coverage because record and reply mechanism cannot work well on multi-core executions. TransFetch shows higher accuracy but lower coverage than Voyager, who performs page prediction that covers wide-range of jumps. 
~\ourwork~achieves both high accuracy and high coverage due to spatio-temporal prefetching. Figure~\ref{fig:ipc} shows the IPC improvement for each application on each input graph.~\ourwork~achieves the highest performance among all the prefetcher: on average the improvement is 12.53\% for GPOP, 21.23\% for X-Stream, and 14.57\% for PowerGraph. Comparing with the top baselines, \ourwork~outperforms TransFetch in GPOP by 3.72\%, Voyager in X-Stream by 4.58\%, and Voyager in PowerGraph by 4.14\%.

\section{Towards Practical Prefetchers}

This paper explores using domain specific ML models to improve memory access prediction and prefetching performance. While these models are not optimized for hardware implementation, we discuss techniques to adapt them for practical use.

\subsection{Reducing Storage Overhead} 

\begin{figure}[t]
  \centering
  \includegraphics[width=1\linewidth]{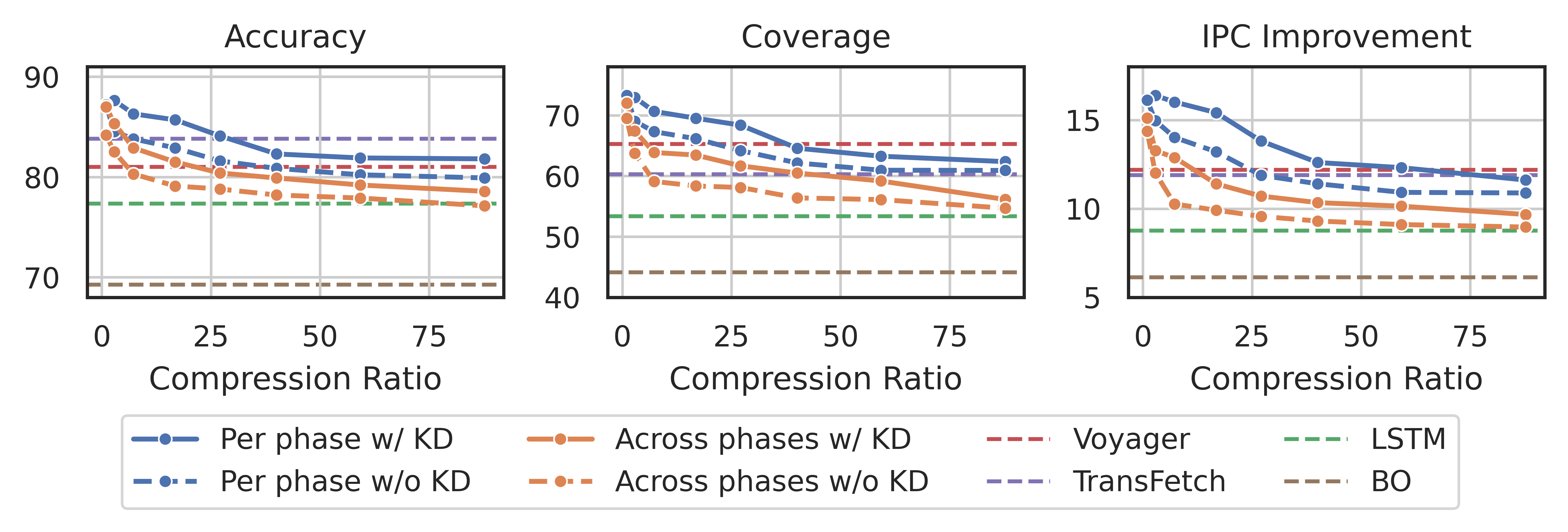}
  \caption{Performance of~\ourwork~under knowledge distillation (KD). Baselines are uncompressed. We compress the models by up to $87\times$ for each phase with 5.47\% higher IPC improvement than BO. 
  }

  \label{fig:kd}
\end{figure}

\begin{figure}[h!]
  \centering
  \includegraphics[width=1\linewidth]{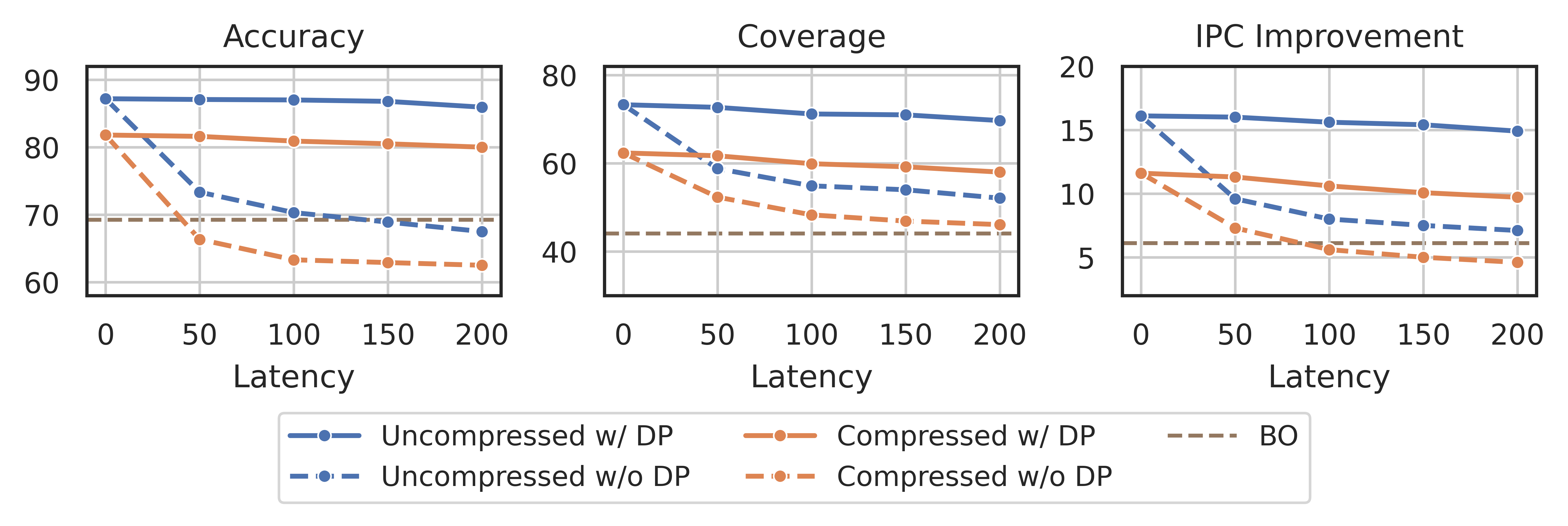}
  \caption{Effectiveness of distance prefetching (DP) for~\ourwork~using the uncompressed and the compressed models. 
  }
  \label{fig:dp}
\end{figure}

{\color{black}
\noindent\textbf{Binary Encoding~\cite{srivastava2019predicting}.} We can use binary-encoding compression to reduce the vocabulary and output dimension of the temporal page predictor. By representing $2^{16}$ classes (page tokens) with a 16-dimension binary vector, we can convert the model output dimension to 16 and the vocabulary of input tokens to 2. This results in up to $33\times$ compression for the model in Table~\ref{tab:model} with $2^{16}$ classes, reducing its parameters from 13M to 397K. The model can be further compressed using knowledge distillation as discussed below.
}
\noindent\textbf{Knowledge Distillation~\cite{hinton2015distilling,gou2021knowledge}.} Knowledge distillation involves transferring knowledge from a larger teacher model to a smaller student model. By tuning the AMMA configuration described in Table~\ref{tab:model} and training them as student models, we reduce the size of the spatial predictor from 419.6K to 7.5K parameters and the temporal predictor from 397K to 1.9K parameters, achieving an overall compression of $87\times$. Figure~\ref{fig:kd} shows that knowledge distillation significantly improves model performance when compressing models. By training a single student model using $N$ phase-specific teacher models, we can further compress the predictor by a factor of $N$ (\# phases per iteration). Although the compression results in performance degradation, ~\ourwork~still shows $5.47\%$ higher IPC improvement compared with the best non-ML prefetcher BO.

\noindent\textbf{Quantization~\cite{polino2018model}.} By representing the weights in the models using 8 bits and applying the above optimizations, we can reduce the storage cost of the spatial model to 7.5KB and the temporal model to 1.9KB. 
 By performing this optimization, the requirement of our model is similar to baseline rule-based prefetchers such as BO (4KB storage) and ISB (8KB storage).

\subsection{Reducing and Hiding Inference Latency}

\noindent\textbf{Parallel Implementation.} {\color{black}Hardware acceleration of neural networks have been widely studied~\cite{geng2021survey,sanaullah2018real}.}
~\ourwork~is based on the attention mechanism, which is highly parallelizable. Using a fully parallel implementation of the AMMA models, we can estimate the latency using the critical path based on Figure~\ref{fig:mm} as follows:
\begin{equation}
T =\underbrace{T_{emb}+T_{att} + T_{fusion} + L \times T_{trans}}_{\text{AMMA}}+\underbrace{T_{hash}}_{\text{Input}}+\underbrace{T_{head}+T_{av}}_{\text{Output}}
\end{equation}
where $T_{emb}$ is for the embedding layer composed of a matrix multiplication (takes $T_{mm}$) and an activation function (takes $T_{av}$), $T_{att}$ is for the self-attention layer that takes $4T_{mm}$ and $3T_{av}$, $T_{fusion}$ is for the multi-modality attention fusion layer that takes $T_{att}+ T_{mm} + 4T_{av}$, $T_{trans}$ is for the Transformer layer with the same critical path with the fusion layer, $T_{hash}$ is for the input processes (hashing, segmentation, and tokenization) that can be implemented in look-up tables with latency 1, and $T_{head}$ is for the output layer that takes $T_{mm}$. Assuming full parallelism, $T_{mm}=1+\log_2D$ for dimension $D$ and $T_{av}=1$ for activation functions using a look-up table, the overall latency is estimated as $T\approx123$ processor cycles for the original model with Transformer dimension $D=128$ and $T\approx79$ for the compressed model $D=8$.

\noindent\textbf{Approximation as a Look-Up Table.}
We can further accelerate the model inference by using Look-Up Tables (LUT) to avoid complex computations.
Recent work has explored approximating matrix multiplication and activation functions using LUT~\cite{razlighi2017looknn}, developing LUT-based Processing-In-Memory approach for fast neural network inference~\cite{ramanathan2020look,besta2021sisa,beneventi2017continuous}, and using LUT for layer-wise approximation~\cite{tang2023lut}. By implementing layer-wise LUT, for models in Figure~\ref{fig:mm} with two sub-layers within the fusion layer and transformer layer~\cite{vaswani2017attention}, the model inference latency can be reduced to approximately 8 cycles regardless of the model dimension.

\noindent\textbf{Distance Prefetching.} 
In addition to reducing the inference latency of a memory access prediction model, distance prefetching~\cite{zhang2022fine} offers an alternative method to hide or offset the latency by skipping the inference slot and predict the future memory accesses in a distance. 
Figure~\ref{fig:dp} shows that distance prefetching  (DP) effectively avoids performance loss caused by model inference latency. With 200 cycles latency introduced in the simulation process, the uncompressed and the $87\times$ compressed~\ourwork~still outperform BO by 8.77\% and 3.58\% w.r.t. IPC improvement, respectively.

{\color{black}
\subsection{Analysis of Computational Complexity}
We use three complexity metrics to demonstrate that our method is state-of-the-art in terms of complexity-performance w.r.t. baseline ML-based prefetchers: 
1) total number of model parameters (Param), 2) number of operations performed in model inference (OPs), and 3) critical path, indicating potential speedups for future parallelized model implementation.
The results are shown in Table~\ref{tab:complexity}. With regard to IPC improvement (IPC Impv), we demonstrate that the MPGraph using 7.2$\times$ compressed models achieves superior performance, utilizing a smaller number of parameters and OPs.
The critical path of our attention-based model (Section~\ref{sec:model},~\cite{vaswani2017attention}) does not depend on the input sequence length $n$ but only depends on the number of layers $l$, 
leading to lower latency compared to LSTM-based models when $n$ is large. 

 \begin{table}[h]
  \caption{{\color{black}Computational Complexity of~\ourwork~and state-of-the-art ML-based prefetchers}}
  \label{tab:complexity}
  \begin{center}
  \begin{tabular}{|c|c|c|c|c|}
    \hline  
    \textbf{ML}&\textbf{Param}&\textbf{OPs}&\textbf{Critical}&\textbf{IPC}\\
    \textbf{Models}&\textbf{(K)}&\textbf{(M)}&\textbf{Path$^\dagger$}&\textbf{Impv(\%)}\\\hline
    Delta-LSTM            & 346.3       & 3.13        & $\mathcal{O}(nl)$         & 8.76          \\\hline
Voyager              & 468.4       & 11.28       &  $\mathcal{O}(nl) $        & 12.2          \\\hline
TransFetch         & 432.5       & 3.38        &  $\mathcal{O}(l)  $        & 11.91         \\\hline
MPGraph             & 837.6       & 7.63        &  $\mathcal{O}(l) $         & 16.11         \\\hline
MPGraph (7.2x) $^\ast$       & 117.6       & 0.99        & $\mathcal{O}(l)  $        & 16.01         \\\hline
    \multicolumn{5}{l}{\small $\dagger$ $n$ is the sequence length; $l$ is the number of layers in the model.} \\
    \multicolumn{5}{l}{\small $\ast$ MPGraph using 7.2$\times$ compressed prediction models.}
\end{tabular}
\end{center}
\end{table}

}

\section{Conclusion}
In this paper, we propose \ourwork, an ML-based prefetcher for graph analytics. Our approach leverages phase, modality, and locality features to develop domain specific models, including phase transition detectors and phase-specific multi-modality models for memory access page and delta prediction. We use a chain spatio-temporal prefetching strategy to manage the models, resulting in 12.53\% to 21.23\% IPC improvement, outperforming state-of-the-art prefetchers. We compress model storage costs via knowledge distillation and reduce inference latency through distance prefetching. The compressed model outperforms non-ML baselines.

DS ML can be extended to many scenarios. For example, Graph frameworks using asynchronous execution allow processes to go beyond the current phase without a barrier for synchronization. The phase transition detector in~\ourwork ~can be extended to each thread for asynchronous frameworks~\cite{low2014graphlab,zhu2015gridgraph,shun2013ligra}. 
DS ML can also be applied to accelerate graph machine learning frameworks~\cite{xia2021graph}. The computation of graph machine learning consists of multiple phases, including sampling, updating and aggregation. Phase detection and phase-specific models can be extended to this domain. Characteristics in graph machine learning such as sparsity in a graph network can also be used in the design of domain specific models.
A general extension of DS ML is to accelerate the training of machine learning models, e.g., neural networks. Each step of optimization consists of iterative phases including fetching batch data, forward-path inference, and back-propagation for weight updates.

\begin{acks}
This work is supported by the DEVCOM Army Research Lab (ARL) under grant W911NF2220159.
\end{acks}

\bibliographystyle{ACM-Reference-Format}
\bibliography{main}

\end{sloppypar}
\end{document}